%% file: main.tex
\PassOptionsToPackage{table}{xcolor}
\documentclass[10pt,letterpaper]{uconn}

\input{math_commands}

\usepackage{amsmath,amssymb}
\usepackage{algorithm}
\usepackage{algpseudocode}
\usepackage{multirow,longtable,array,tabularx}
\usepackage{textcomp,wrapfig,needspace,flafter,enumitem,titletoc}
\input{visual_style}

\definecolor{algprivate}{HTML}{FFEEDD}
\colorlet{algagent}{anttint}
\makeatletter
\newcommand{\algrow}[2]{%
  {\setlength{\fboxsep}{0pt}%
   \hspace*{-\ALG@thistlm}%
   \colorbox{#1}{\makebox[\dimexpr\linewidth-0.71pt\relax][l]{\strut\hspace*{\ALG@thistlm}#2}}}%
}
\makeatother
\newcommand{\algchip}[2]{{\setlength{\fboxsep}{1.2pt}\colorbox{#1}{#2}}}
\newtcolorbox{swespecpanel}[1][]{%
  enhanced,frame hidden,boxrule=0pt,
  borderline west={1.1pt}{0pt}{antprimarydark},
  colback=antprompt,sharp corners,boxsep=0pt,
  left=8pt,right=8pt,top=6pt,bottom=6pt,
  before skip=5pt,after skip=7pt,#1
}

\title{SWE-Trail: History-Guided Trajectory 
  Collection for \\ Coding Agent Training}
\author{%
  Ziming Li\textsuperscript{1,2},
  Lu Yu\textsuperscript{2},
  Jiatan Huang\textsuperscript{1},
  Qing Cui\textsuperscript{2},
  Jun Zhou\textsuperscript{2,\dag},
  Chuxu Zhang\textsuperscript{1,\dag}}
\affiliation{%
  \textsuperscript{1}University of Connecticut\quad
  \textsuperscript{2}Ant Group\\[2pt]
  \texttt{\{ziming.li,chuxu.zhang\}@uconn.edu}\enspace
  \texttt{jun.zhoujun@antgroup.com}\enspace
  \textsuperscript{\dag}Corresponding authors}
\runningtitle{SWE-Trail}
\hypersetup{
  pdftitle={SWE-Trail: History-Guided Trajectory Collection for Coding Agent Training},
  pdfauthor={Ziming Li, Lu Yu, Jiatan Huang, Qing Cui, Jun Zhou, Chuxu Zhang},
  pdfsubject={History-guided trajectory collection for coding agent training},
  pdfkeywords={SWE-Trail, coding agents, training data, trajectory collection}
}
\input{sections/00_abstract}

\begin{document}
\maketitle
\thispagestyle{firstpage}

\input{sections/01_introduction}
\input{sections/03_motivating_study}
\input{sections/04_method}
\input{sections/05_experiments}
\input{sections/06_results}
\input{sections/07_related_work}
\input{sections/08_discussion}

\input{sections/09_conclusion}

\bibliographystyle{plainnat}
\bibliography{references}

\clearpage
\appendix
\input{appendix/appendix}

\end{document}

%% file: math_commands.tex
\usepackage{amsmath,amsfonts,bm}

\def\eqref#1{equation~\ref{#1}}

\def\1{\bm{1}}

\DeclareMathAlphabet{\mathsfit}{\encodingdefault}{\sfdefault}{m}{sl}
\SetMathAlphabet{\mathsfit}{bold}{\encodingdefault}{\sfdefault}{bx}{n}



%% file: visual_style.tex
\definecolor{antprimary}{HTML}{1677FF}
\definecolor{antprimarydark}{HTML}{0958D9}
\definecolor{antdeep}{HTML}{003EB3}
\definecolor{antlight}{HTML}{69B1FF}
\definecolor{anttint}{HTML}{E6F4FF}
\definecolor{antprompt}{HTML}{F7FBFF}
\definecolor{baselinegray}{HTML}{8C8C8C}
\definecolor{baselinegraydark}{HTML}{595959}
\definecolor{controlledgray}{HTML}{BFBFBF}
\definecolor{controlledlight}{HTML}{D9D9D9}
\definecolor{slate}{HTML}{4A5568}
\definecolor{text}{HTML}{262626}
\definecolor{muted}{HTML}{595959}
\definecolor{grid}{HTML}{E8E8E8}
\definecolor{axis}{HTML}{8C8C8C}
\definecolor{panel}{HTML}{FFFFFF}
\definecolor{background}{HTML}{FFFFFF}
\definecolor{warning}{HTML}{D46B08}
\newcommand{\mainrow}{\rowcolor{anttint}}

\newcommand{\tablegain}[1]{\textcolor{antprimarydark}{\scriptsize\thinspace(#1)}}
\newcommand{\tabledrop}[1]{\textcolor{baselinegraydark}{\scriptsize\thinspace(#1)}}

\tcbset{swecardbase/.style={enhanced,boxrule=0.8pt,arc=0pt,left=7pt,right=7pt,top=6pt,bottom=6pt,before skip=5pt,after skip=7pt}}
\newtcolorbox{sweprompt}[1]{swecardbase,breakable,title={#1},colframe=antprimarydark,colback=antprompt,colbacktitle=antprimarydark,coltitle=white,fonttitle=\bfseries\small,fontupper=\ttfamily\footnotesize}
\newtcolorbox{swecasecard}[2][]{swecardbase,title={#2},colframe=antprimarydark,colback=panel,colbacktitle=antprimarydark,coltitle=white,fonttitle=\bfseries\scriptsize,fontupper=\scriptsize,before skip=0pt,after skip=0pt,#1}
\newtcolorbox{swedossier}[2][]{enhanced,breakable,boxrule=0.7pt,arc=1pt,colframe=antlight,colback=panel,left=8pt,right=8pt,top=9pt,bottom=8pt,title={#2},attach boxed title to top left={xshift=8pt,yshift=-2.4mm},boxed title style={boxrule=0pt,arc=1pt,colback=antprimarydark,colframe=antprimarydark},coltitle=white,fonttitle=\bfseries\small,before skip=9pt,after skip=9pt,#1}
\newtcolorbox{sweartifact}[1]{enhanced,breakable,boxrule=0.5pt,arc=1pt,colframe=grid,colback=antprompt,left=7pt,right=7pt,top=7pt,bottom=7pt,title={#1},attach boxed title to top right={xshift=-5pt,yshift=-2mm},boxed title style={boxrule=0pt,arc=1pt,colback=anttint,colframe=anttint},coltitle=antprimarydark,fonttitle=\bfseries\scriptsize,fontupper=\ttfamily\scriptsize,before skip=7pt,after skip=5pt}

%% file: sections/00_abstract.tex
\paperabstract{
Coding agents solve repository issues through long sequences of exploration, editing, and validation. These interaction trajectories are increasingly used as post-training data, so how they are collected shapes what the trained model learns. A common approach is to sample multiple trajectories independently for each issue and then select or curate them. 
However, as more trajectories are sampled independently, later trajectories become increasingly redundant. In fact, earlier trajectories can guide where subsequent collection effort should be spent. A natural way to use this history is to expose it to the coding agent as advice, helping it decide what to do next. But training on such trajectories can make the model rely on advice that is unavailable at test time, creating a train-test mismatch that may reduce downstream performance. Instead, we argue that history should guide \emph{where} collection effort is spent, while the coding agent decides \emph{what} to generate by itself.
Based on this principle, we introduce \textbf{SWE-Trail}, a trajectory collector that uses earlier trajectories to guide subsequent collection while keeping that history outside the coding agent's context. During a new rollout, SWE-Trail uses history to identify a state where additional collection effort may be useful. When that state is reached, the coding agent proposes candidate next actions from its ordinary context. SWE-Trail briefly executes candidates from the same checkpoint and uses the results to decide which route continues. We compare collection approaches by training student models on the trajectories they produce and measuring downstream performance. For a 124B student initialized from Ling-3.0-flash-base, SWE-Trail improves mean resolution over independent sampling by 2.33 and 1.60 percentage points on SWE-bench Multilingual and SWE-bench Pro, respectively. Without history, the same collection pipeline produces more resolved trajectories than independent sampling, but improves by only 0.11 points on Multilingual. These results show the value of using history to guide collection without exposing it to the coding agent.
}

%% file: sections/01_introduction.tex
\section{Introduction}
\label{sec:introduction}
Resolving a repository-level issue requires a coding agent to explore the repository, modify code, and validate its changes~\citep{jimenez2023swebench,yang2024sweagent,wang2024openhands}. These interactions form trajectories that record not only the final patch but also the agent's reasoning and actions throughout the repair process. Such trajectories are increasingly used as post-training data, so the collection process directly affects their value for training~\citep{chen2026tdscaling,han2026trajectorycuration}.

A common way to collect coding agent training trajectories is to sample several trajectories per issue and keep those that pass tests or a judge~\citep{pan2024swegym,ehrlich2025codemonkeys,jain2025r2egym}. Some work also edits the kept trajectories, masking flawed tool calls or removing redundant steps~\citep{tao2026swelego,ma2026patches,han2026swetrace,xiao2025agentdiet}. These methods operate after trajectories are collected, while trajectories for the same issue are still generated independently. However, later independent samples become increasingly redundant as more trajectories are collected. They often revisit repair targets and behaviors already explored earlier. Rather than ignoring this history, subsequent collection can use earlier trajectories to decide where additional collection effort should be spent. For example, disagreements between successful and failed trajectories can reveal decision points where alternative actions are worth comparing. We refer to such evidence as \emph{allocation signals} and to the earlier trajectories for an issue as its \emph{history}. Section~\ref{sec:motivation} quantifies the redundancy of independent sampling and tests whether earlier trajectories provide such allocation signals. Since earlier trajectories can provide useful information for a new rollout, a natural way to use this history is to place it in the coding agent's context as advice~\citep{chen2025sweexp,wang2026memgovern}. This can help the current rollout, but the resulting training trajectories record reasoning and actions generated with access to the advice. However, at test time, the trained model must make decisions without that advice. As a result, these trajectories may provide less useful supervision for training, as we evaluate in Section~\ref{sec:student-results}. This leads to our central question:
\textbf{\emph{How can history guide where collection effort is spent while the coding agent decides what to generate from its ordinary context?}}

To answer this question, we introduce \textbf{SWE-Trail}, a trajectory collector that keeps history outside the coding agent's context. Before collection, SWE-Trail uses history to determine which states should receive more collection effort and what evidence should be considered at those states. During a new rollout, when the agent reaches such a state, it proposes candidate next actions from its ordinary context without access to the history. SWE-Trail then compares candidate actions by briefly executing them from the same checkpoint. It selects the route that is better supported by repository evidence. The coding agent then continues from the selected route to complete the trajectory. A concrete example illustrates this process. On a Rust issue, all prior failed trajectories made an edit and then continued reading without running tests. From this history, SWE-Trail determines that this state should receive additional collection effort when it appears again in a new rollout. When the rollout reaches such a state, the coding agent proposes different next actions from its ordinary context, including further reading and running tests. SWE-Trail briefly executes both choices from the same checkpoint and selects the test-running route based on the resulting evidence. The coding agent then continues from the selected route until the rollout ends.
We evaluate whether SWE-Trail collects trajectories that are more useful for downstream training. The comparison uses the same 300 SWE-rebench V2 issues~\citep{badertdinov2026swerebenchv2}. Each setting collects five trajectories per issue with GLM-5.1~\citep{glm5team2026glm5,zai2026glm51}. \emph{SeedBank} provides the independent sampling baseline. \emph{H5} uses the five SeedBank trajectories for each issue as history to collect five new trajectories, while \emph{H0} runs the same collection pipeline without history. We train the 124B Ling-3.0-flash-base model~\citep{inclusionai2026ling3flashbase} on the trajectories from each condition, with three independent training runs. Under this controlled setup, H5 improves mean resolution over SeedBank by 2.33 and 1.60 percentage points on SWE-bench Multilingual and SWE-bench Pro, respectively. The H0 control helps isolate the contribution of history. It produces more resolved trajectories than SeedBank, but improves the 124B student's mean resolution by only 0.11 points on Multilingual. This contrast shows that collection success alone does not explain the downstream gain. A second student model, Qwen3-14B-Base~\citep{qwenteam2025qwen3base}, is also trained on the same H5 trajectories to test whether the gain generalizes across student models. For Qwen3-14B-Base, H5 improves mean resolution over SeedBank by 3.11 and 2.42 percentage points on SWE-bench Multilingual and SWE-bench Pro, respectively. Together, these results support using history to guide trajectory collection without placing that history in the coding agent's context.

%% file: sections/03_motivating_study.tex
\section{Motivation}
\label{sec:motivation}
To examine the empirical basis for our motivation, we analyze the independently sampled trajectories in SeedBank. 
We use \emph{attempt} to refer to one independent try at solving an issue, and examine both how coverage grows with additional attempts and whether earlier attempts provide allocation signals for subsequent collection.
\subsection{Diminishing Coverage Gains from Independent Rollouts}
\label{sec:marginal-coverage}

\begin{figure}[t]
	\centering
	\includegraphics[width=\textwidth]{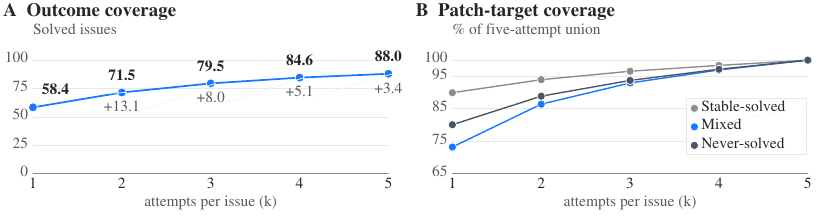}
	\caption{\textbf{Diminishing coverage gains from independent rollouts.}
		A: expected number of issues solved with $k$ attempts per issue.
		Gray labels show marginal additions.
		B: mean coverage of each issue's five-attempt patch-target union,
		grouped by SeedBank outcomes.
		Both panels average over all $k$-attempt subsets of the five recorded attempts.
		The vertical axis in B starts at 65\%.}
	\label{fig:motivation-gap}
\end{figure}

Figure~\ref{fig:motivation-gap}A shows the expected number of issues solved with $k$ attempts per issue, averaging over all $k$-attempt subsets of the five SeedBank attempts for each issue.
For a given $k$-attempt subset, an issue is counted as solved if at least one of the $k$ attempts succeeds.
With one attempt per issue, 58.4 issues are solved in expectation. Increasing $k$ from 1 to 2 adds 13.1 solved issues, while increasing it to 3, 4, and 5 adds only 8.0, 5.1, and 3.4 more issues, respectively. Thus, each additional attempt still expands issue coverage, but by a smaller amount. 

Patch-target coverage shows the same diminishing pattern across different types of history.
We use modified files as patch targets and define three history types from each issue's SeedBank outcomes: \emph{stable-solved} if all attempts succeed, \emph{mixed} if both successful and failed attempts occur, and \emph{never-solved} if all attempts fail.
For each history type, Figure~\ref{fig:motivation-gap}B shows the fraction of files modified across an issue's five attempts that are covered by $k$ attempts. We average this fraction over all $k$-attempt subsets and then across issues. 
One attempt already covers 90.0\% of these files on stable-solved issues, 73.2\% on mixed issues, and 80.1\% on never-solved issues. These results show substantial target overlap across all three history types. As more attempts are added, later attempts cover fewer new files. Overlap among successful attempts may simply reflect convergence on the same repair targets. However, even when all independent attempts fail, they still modify many of the same files. This pattern shows that independent sampling does not automatically redirect collection effort after repeated failures. Later attempts may continue spending effort on previously explored repair targets rather than exploring new ones. Together with the issue-level results, these findings show that later independent attempts become increasingly redundant.

\subsection{History Provides Allocation Signals}
\label{sec:mixed-boundaries}
%

Section~\ref{sec:marginal-coverage} shows that later independent attempts add less coverage and often revisit repair targets explored earlier.
The remaining question is whether history can indicate where additional collection effort may be useful.
Mixed issues are informative because successful and failed attempts may choose different next actions from similar states.
We therefore examine whether such decision points occur.

For each mixed issue, an auditor compares the most similar successful and failed attempts and locates their first meaningful decision difference. We assess whether the two attempts are similar enough to compare their next actions and whether both actions are reasonable at that point. The audit is conducted independently with GPT-5.6-Sol~\citep{openai2026gpt56} and GLM-5.1. The main analysis includes 43 of the 51 mixed issues, excluding eight whose failed attempt ended in a timeout or collector error. GPT-5.6-Sol identifies such cases in 26 issues and GLM-5.1 in 31, with 20 identified by both models. These results show that mixed histories can reveal decision points where additional collection effort may be useful. The detailed results are reported in Appendix~\ref{app:cross-model-route-audit}. 

However, this evidence relies on having both successful and failed attempts. 
To test whether history remains useful without one, we examine never-solved histories and measure the overlap in modified files across failed attempts on the same issue. 
The mean pairwise Jaccard overlap is 0.841. Among the 198 issues with a defined overlap score, 93 have a score of 1.0. These results show that failed attempts often focus on the same repair targets. This overlap does not explain why the attempts fail, but it shows where effort is repeatedly spent across failed attempts. Thus, even without a success--failure contrast, history can still indicate where additional collection effort may be useful. Together, these findings show that history can provide allocation signals for collection.
Appendix~\ref{app:motivation-analyses} gives the audit procedure and overlap calculations. 

\begin{figure}[t]
	\centering
	\includegraphics[width=\textwidth]{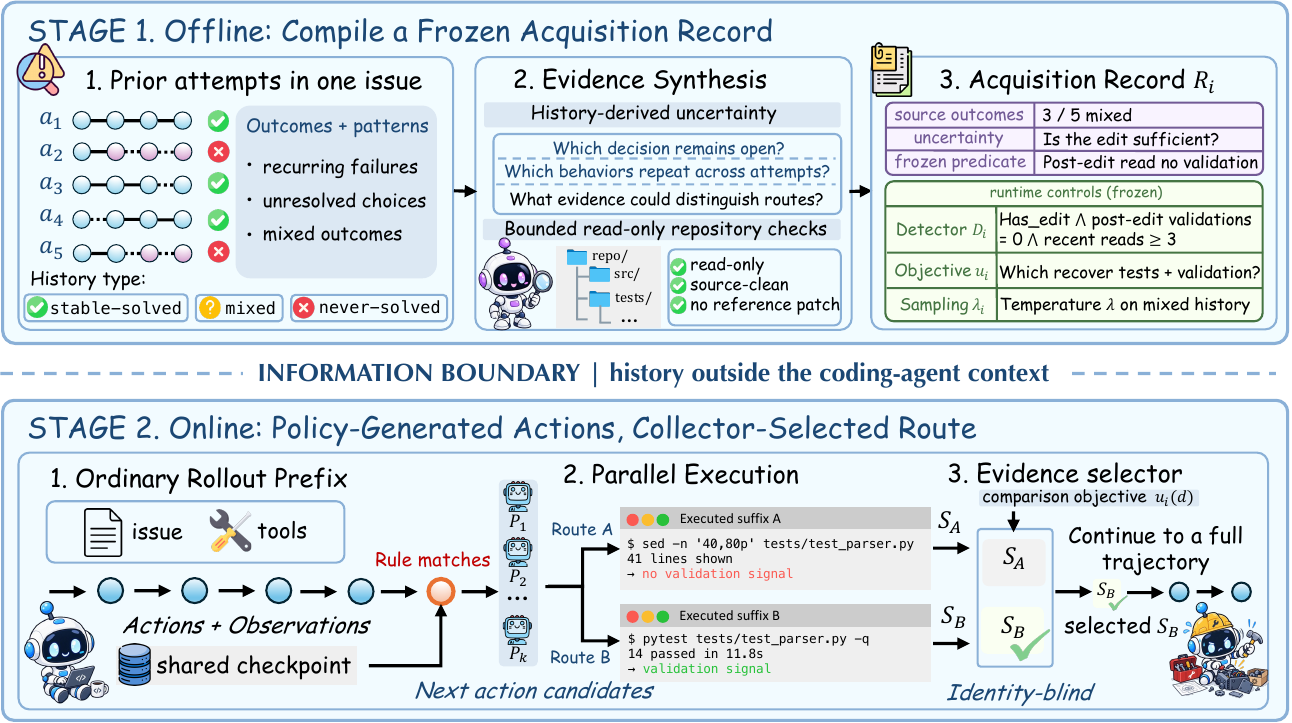}
	\caption{\textbf{SWE-Trail in two stages.} Offline, the compiler reads the issue history, runs bounded read-only repository checks, and freezes an acquisition record $R_i$ containing a detector $D_i$, a comparison objective $u_i$, and a sampling schedule $\lambda_i$. Online, when $D_i$ identifies a state for additional collection effort, the coding agent proposes candidate next actions from its ordinary context. SWE-Trail briefly executes two candidates from the same checkpoint, and an identity-blind selector retains the route better supported by repository evidence under $u_i$. The coding agent then continues from the selected route. History never enters the coding agent's context.}
	\label{fig:method-overview}
\end{figure}
%
%

%% file: sections/04_method.tex
\section{SWE-Trail}
\label{sec:method}

Section~\ref{sec:motivation} shows that later independent attempts add less new coverage, while earlier attempts can indicate where additional collection effort may be useful.
Motivated by these observations, we introduce SWE-Trail, a trajectory collector that uses history to decide where additional collection effort should be spent in a new rollout. 
For issue $i$, let $x_i$ contain the issue statement and the repository snapshot at the issue commit. The history $H_i$ contains the attempts already collected on that issue. SWE-Trail has access to $H_i$, but the coding agent does not. Reference patches are unavailable during collection. Figure~\ref{fig:method-overview} shows the two stages of SWE-Trail. Before collection, SWE-Trail uses the history and read-only repository checks to determine which states should receive additional collection effort and what evidence should be considered at those states. During a rollout, when such a state is reached, the coding agent proposes candidate next actions from its ordinary context without access to the history. SWE-Trail selects candidates for comparison, briefly executes them in parallel from the same checkpoint, and retains the route better supported by repository evidence. The coding agent then continues from the selected route to complete the trajectory.

\subsection{Compiling History into Detectors}
\label{sec:branch-opportunity}

Earlier attempts can indicate which states should receive additional collection effort, but a new rollout may reach similar situations through a different sequence of actions. The compiler therefore builds a detector $D_i$ whose rules recognize such states from the rollout's actions and observations rather than relying on fixed step numbers. For example, the detector in Figure~\ref{fig:method-overview} recognizes repeated reading after an edit without validation. The runtime checks these rules after each new repository observation and samples candidate actions when a state matches.

To build the detector, the compiler first analyzes $(x_i,H_i)$ to extract allocation signals from the history. These signals include decisions supported by successful attempts, differences across repair routes, and recurring failures. The history may leave open questions about the issue or repository, so the compiler then runs bounded read-only repository checks to resolve these questions. These checks do not modify source code or access the reference patch. The compiler stores the historical evidence, check results, and remaining questions in a private acquisition record $R_i$. Using $R_i$, the compiler defines the detector rules and a local comparison objective $u_i(d)$ for each detector state $d$. The objective specifies what evidence should be considered when comparing candidate routes at that state. Different histories lead the compiler to focus on different kinds of decisions. For stable-solved histories, repeated success shows which decisions already have strong support. The compiler allocates additional effort only when meaningful uncertainty remains. For mixed histories, it focuses on states where successful and failed repair routes differ. For never-solved histories, it focuses on recurring failure patterns, such as edits followed by further reading without validation. 
The compiler also sets a sampling-temperature schedule $\lambda_i$ for candidate generation when a detector state is reached. Mixed and never-solved histories use higher temperatures to encourage more varied next actions when the history shows unresolved choices or repeated failures.

Before collection begins, the compiler validates each proposed detector state against its supporting evidence and replays its rules on the recorded attempts. States without sufficient support are discarded. The admitted detector states, their comparison objectives, and $\lambda_i$ are then fixed for the entire collection round. Appendices~\ref{app:offline-compiler} and~\ref{app:compiler-records} provide the compiler details, admission checks, and example records.

\subsection{Sampling Comparable Alternatives}
\label{sec:online-acquisition}

The detector determines where additional collection effort is spent, while the actions to consider come from the coding agent itself. When the rollout reaches a state $d$ recognized by $D_i$, SWE-Trail samples several candidate next actions using only the agent's ordinary context. The history, acquisition record $R_i$, and comparison objective $u_i(d)$ are not added to that context. The first sample uses the agent's ordinary temperature and serves as the default. The remaining samples use $\lambda_i$.

Different samples may propose equivalent actions, so SWE-Trail first groups equivalent proposals together. It keeps the default candidate and shortlists executable alternatives from the remaining proposals. To keep each intervention local and bounded, SWE-Trail chooses at most one alternative from this shortlist using $R_i$ and $u_i(d)$. If no suitable alternative is available, the rollout continues with the default action. Otherwise, SWE-Trail restores two isolated workspaces from the same checkpoint, one for the default and one for the chosen alternative. Both candidates start from the same conversation and workspace state and run for the same short horizon, producing two \emph{executed suffixes} of actions and observations. To avoid expanding collection into a multi-level search tree, each trajectory performs at most one such comparison. Different rollouts on the same issue may still make their comparison at different detector states. 
\subsection{Selecting a Route by Executed Evidence}
\label{sec:evidence-selection}

Once both suffixes have been executed, SWE-Trail compares their progress using the local comparison objective $u_i(d)$. The two suffixes are presented to SWE-Trail in shuffled order, hiding which one comes from the default candidate. SWE-Trail uses $R_i$ only as background for the comparison. It prefers a route only when the two executed suffixes provide clear evidence that one makes better progress under $u_i(d)$. Otherwise, it keeps the default. After selection, the coding agent continues only from the selected workspace until the full rollout ends. For training, SWE-Trail keeps only the selected route, preserves the coding agent's original reasoning and actions without rewriting, and removes the unselected suffix and private collector events. This ensures that the training record remains a single continuous trajectory generated by the coding agent and that private collection information does not enter the training process. Algorithm~\ref{alg:swe-trails-full} in Appendix~\ref{app:offline-compiler} summarizes the procedure, and Appendix~\ref{app:rescue-case} provides a full end-to-end case study.

%% file: sections/05_experiments.tex
\begin{table}[!t]
	\centering
	\setlength{\abovecaptionskip}{0pt}
	\setlength{\belowcaptionskip}{5pt}
	\caption{\textbf{H5 improves downstream resolution on both benchmarks.} Mean resolution over three independent training runs. Changes are relative to SeedBank, and ranges show the lowest and highest result. Rows named after prior work use the same SeedBank attempts as controlled adaptations. Table~\ref{tab:condition-matrix-full} defines all conditions.}
	\label{tab:main-sft-results}
	\footnotesize
	\setlength{\tabcolsep}{2pt}
	\resizebox{\textwidth}{!}{%
		\begin{tabular}{@{}lrlcrlc@{}}
			\toprule
			& \multicolumn{3}{c}{Multilingual} & \multicolumn{3}{c}{SWE-Pro} \\
			\cmidrule(lr){2-4}\cmidrule(l){5-7}
			Coding-trajectory condition & \multicolumn{2}{c}{Mean} & Range & \multicolumn{2}{c}{Mean} & Range \\
			\midrule
			SeedBank independent rollouts & 63.33 & & {\scriptsize [63.00, 63.67]} & 41.22 & & {\scriptsize [41.18, 41.31]} \\
			\addlinespace[2pt]
			SWE-Gym~\citep{pan2024swegym}, {\scriptsize outcome-conditioned transform} & 64.00 & \tablegain{+0.67} & {\scriptsize [63.33, 65.00]} & 40.86 & \tabledrop{\textminus0.36} & {\scriptsize [40.36, 41.59]} \\
			SWE-TRACE~\citep{han2026swetrace}, {\scriptsize critical-path reduction} & 64.44 & \tablegain{+1.11} & {\scriptsize [63.33, 65.33]} & 40.77 & \tabledrop{\textminus0.46} & {\scriptsize [40.08, 41.72]} \\
			SWE-Lego~\citep{tao2026swelego}, {\scriptsize strict tool-error masking} & 64.56 & \tablegain{+1.22} & {\scriptsize [63.00, 66.00]} & 41.63 & \tablegain{+0.41} & {\scriptsize [39.81, 42.95]} \\
			SWE-Exp~\citep{chen2025sweexp}, {\scriptsize agent-visible experience} & 62.00 & \tabledrop{\textminus1.33} & {\scriptsize [61.00, 63.00]} & 38.44 & \tabledrop{\textminus2.78} & {\scriptsize [38.30, 38.58]} \\
			\midrule
			\textsc{SWE-Trail-H0} & 63.44 & \tablegain{+0.11} & {\scriptsize [63.00, 63.67]} & 41.95 & \tablegain{+0.73} & {\scriptsize [41.04, 43.09]} \\
			\textsc{SWE-Trail-H5} (random detection) & 64.22 & \tablegain{+0.89} & {\scriptsize [63.33, 64.67]} & 41.54 & \tablegain{+0.32} & {\scriptsize [41.31, 41.72]} \\
			\textsc{SWE-Trail-H5} (random retention) & 64.78 & \tablegain{+1.44} & {\scriptsize [63.67, 66.00]} & 41.63 & \tablegain{+0.41} & {\scriptsize [41.04, 42.13]} \\
			\mainrow \textbf{\textsc{SWE-Trail-H5}} & \textbf{65.67} & \tablegain{+2.33} & {\scriptsize [64.00, 67.67]} & \textbf{42.82} & \tablegain{+1.60} & {\scriptsize [41.72, 43.37]} \\
			\bottomrule
	\end{tabular}}
\end{table}

\vspace{0.1in}
\section{Experiments}
\label{sec:results}
\subsection{Setting}
\label{sec:experiments}
We test whether training on SWE-Trail trajectories improves student performance on new issues.
Across all comparisons, we use the same collection issues and the same number of coding trajectories for training.
The background training data and the training and evaluation protocols remain fixed.

\textbf{Shared setting.}
Each training dataset contains five coding trajectories for each of the same 300 SWE-rebench~V2 issues~\citep{badertdinov2026swerebenchv2}, giving 1,500 trajectories in total. None of the collection repositories appears in either evaluation benchmark. All coding rollouts use GLM-5.1~\citep{glm5team2026glm5} with the OpenHands scaffold~\citep{wang2024openhands}. Collection runs in an offline sandbox with repository commit histories removed. Reference patches and published fixes are unavailable during collection. Appendix~\ref{app:data} details the provenance of the collection issues and verifies that they share no issues or repositories with either evaluation benchmark.

\textbf{Methods and variants.}
SeedBank provides the independent-sampling baseline. We apply controlled adaptations of SWE-Gym~\citep{pan2024swegym}, SWE-TRACE~\citep{han2026swetrace}, and SWE-Lego~\citep{tao2026swelego} to the same SeedBank trajectories. We also adapt SWE-Exp~\citep{chen2025sweexp} to test a different use of history, where the coding agent reads an experience message compiled from the SeedBank attempts during collection. The message is retained in the training record. 

For SWE-Trail, history remains outside the coding agent's context throughout collection. Its main variant, H5, uses the five SeedBank attempts for each issue as history when collecting five new trajectories. H0 sets $H_i=\varnothing$ and runs the same read-only repository checks and compiler without history. Comparing H0 with SeedBank tests the collection pipeline without history, while comparing H5 with H0 isolates what history adds. We also use two H5 variants to test where SWE-Trail makes comparisons and which route it retains. \emph{Random detection} replaces the history-guided comparison location with a random eligible state while keeping H5's remaining collection procedure unchanged. \emph{Random retention} keeps H5's detector but randomly chooses which executed route to continue. Both variants use the same per-trajectory model-call limits and execution horizon as H5.

\begin{table}[t]
	\centering
	\setlength{\abovecaptionskip}{0pt}
	\setlength{\belowcaptionskip}{5pt}
	\caption{\textbf{H10 costs 12.9\% more recorded tokens than H5 for a 0.22/0.14 point gain.} Differences are relative to H0. Acquisition-only tokens exclude the ordinary rollout and total tokens include it. $\dagger$: H0 omits offline compiler calls, so its compute is a lower bound and the differences are upper bounds.}
	\label{tab:history-cost}
	\label{tab:cost-results}
	\footnotesize
	\setlength{\tabcolsep}{3pt}
	\resizebox{\textwidth}{!}{%
		\begin{tabular}{@{}lrrrrrrr@{}}
			\toprule
			& & \multicolumn{2}{c}{Acquisition} & \multicolumn{2}{c}{Recorded compute} & \multicolumn{2}{c}{Student utility} \\
			\cmidrule(lr){3-4}\cmidrule(lr){5-6}\cmidrule(l){7-8}
			Condition & Hist. & Res. & Issues & Acq.-only & Total & Multi. & Pro \\
			\midrule
			\textsc{SWE-Trail-H0} & 0 & 329 & 92 & 164M$^{\dagger}$ & $\geq$1.845B$^{\dagger}$ & 63.44 & 41.95 \\
			\mainrow \textbf{\textsc{SWE-Trail-H5} (main)} & 5 & \textbf{345}\tablegain{+16} & \textbf{98}\tablegain{+6} & 307M\tablegain{+143M$^{\dagger}$} & 1.914B\tablegain{+69M$^{\dagger}$} & 65.67\tablegain{+2.22} & 42.82\tablegain{+0.87} \\
			\textsc{SWE-Trail-H10} & 10 & 341\tablegain{+12} & 94\tablegain{+2} & 510M\tablegain{+346M$^{\dagger}$} & 2.160B\tablegain{+315M$^{\dagger}$} & \textbf{65.89}\tablegain{+2.44} & \textbf{42.95}\tablegain{+1.00} \\
			\bottomrule
	\end{tabular}}
\end{table}
To examine history scale, H10 adds five independently sampled attempts to the five attempts already used by H5 for each issue. It otherwise uses the same read-only repository checks and compiler, so comparing H10 with H5 tests whether more history adds value. We also compare SWE-Trail with verifier-based trajectory selection. \emph{Top-5-of-10} selects five of ten independently sampled trajectories per issue. \emph{Top-5-hybrid} applies the same verifier to five SeedBank and five H5 trajectories and selects five from the combined set. Table~\ref{tab:condition-matrix-full} in Appendix~\ref{app:data} gives the full experimental conditions. Appendices~\ref{app:online-runtime} and~\ref{app:contracts} provide the runtime settings and export checks.

\textbf{Students and evaluation.}
Every trajectory in the final training dataset is used, whether resolved or not. The primary student starts from the 124B Ling-3.0-flash-base model~\citep{inclusionai2026ling3flashbase} and is trained with full-parameter supervised fine-tuning on 128 H800 GPUs. Each dataset is used for three independent training runs. Evaluation uses SWE-bench Multilingual~\citep{yang2025swesmith} with 300 tasks and SWE-bench Pro~\citep{deng2025swebenchpro} with 731 tasks. We refer to the latter as SWE-Pro below. Resolution is the fraction of tasks whose patches pass the benchmark tests, and the reported score is the mean across the three training runs. A second student, Qwen3-14B-Base~\citep{qwenteam2025qwen3base}, is trained to test whether the gains extend to another model. Appendix~\ref{app:training} gives the training and evaluation protocols and metric definitions.

%% file: sections/06_results.tex
\subsection{Downstream Training Utility}
\label{sec:student-results}
\textbf{SWE-Trail improves downstream resolution.}
Table~\ref{tab:main-sft-results} shows that H5 outperforms all other conditions in the table on both benchmarks. H5 improves mean resolution over SeedBank by 2.33 percentage points on Multilingual and 1.60 points on SWE-Pro. Since the dataset size and training setup are fixed, these differences reflect changes in the collected trajectories. Across three independent training runs, every H5 run exceeds every SeedBank run on both benchmarks. An exact one-sided permutation test over the training runs gives $p=0.05$ on each benchmark. Appendix~\ref{app:downstream-results} reports the individual runs, statistical tests, and results by language.
SWE-Exp falls below SeedBank by 1.33 points on Multilingual and 2.78 points on SWE-Pro. This may reflect reliance on advice available during training but absent at evaluation. To examine the role of advice more directly, Appendix~\ref{app:advice-exports} compares two exports of the same trajectories, with the advice either retained or removed. Compared with retaining the advice, removing it improves SWE-Pro but lowers Multilingual. Removing the advice after collection does not remove its influence on how the trajectory was generated. The agent produced its reasoning after reading the advice, so some decisions may remain without the information that originally supported them.

\textbf{H5 has higher mean resolution than H0.}
Without history, H0 improves mean resolution over SeedBank by 0.11 points on Multilingual and 0.73 points on SWE-Pro. Adding history in H5 raises mean resolution by a further 2.22 and 0.87 points over H0, respectively. The point estimates are positive on both benchmarks, but task-paired bootstrap gives 95\% intervals of $[-0.2, 4.6]$ points on Multilingual and $[-0.8, 2.5]$ on SWE-Pro.  Appendix~\ref{app:paired-contrasts} gives the full paired analysis. Section~\ref{sec:corpus-results} compares these downstream differences with the corresponding collection outcomes.

\textbf{Both collection decisions matter.}
To test these two choices, we randomize each one separately.
\emph{Random detection} replaces H5's history-guided choice of where to compare with a random eligible point and lowers mean resolution by 1.44 points on Multilingual and 1.28 points on SWE-Pro.
\emph{Random retention} keeps H5's history-guided comparison location but randomly chooses which executed route continues, lowering mean resolution by 0.89 and 1.19 points, respectively.
These decreases support using history to decide where additional collection effort should be spent and execution evidence to decide which route should continue.
Both variants remain above SeedBank in mean resolution, while their run ranges overlap H5's on both benchmarks.

\textbf{The gains extend to another student.} We repeat the comparison with Qwen3-14B-Base using the same SeedBank, H0, and H5 trajectory sets. H5 improves mean resolution over SeedBank by 3.11 percentage points on Multilingual and 2.42 points on SWE-Pro. It also exceeds H0 by 1.67 and 1.60 points, respectively. These results show that the downstream gains from SWE-Trail also appear with another student model. Appendix~\ref{app:second-student} gives the full results.

\vspace{-0.05in}
\subsection{History Scale and Cost}
\label{sec:cost-results}
\vspace{-0.05in}

\textbf{More history brings smaller gains.} Table~\ref{tab:history-cost} compares H0, H5, and H10. Using five prior attempts improves mean resolution over H0 by 2.22 points on Multilingual and 0.87 points on SWE-Pro. Increasing the history from five to ten attempts adds only 0.22 and 0.14 points, respectively. H10 also does not improve the collection outcomes over H5, despite increasing tokens spent beyond the ordinary rollout from 307M to 510M, a 66.0\% increase, and total collection tokens by 12.9\%. More history therefore provides only small additional mean gains at substantially higher collection overhead. The main H5 setting remains close to H0 in cost. Excluding the cost of collecting prior attempts, H5 uses at most 3.7\% more tokens than H0. Its ordinary rollouts use 4.4\% fewer tokens than H0. The same diminishing gain appears with Qwen3-14B-Base. H5 improves over H0 by 1.67 and 1.60 points, while H10 adds a further 1.00 and 0.36 points on Multilingual and SWE-Pro, respectively. Appendix~\ref{app:cost} gives the complete cost breakdowns.

\Needspace{12\baselineskip}
\begin{wraptable}{r}{0.46\textwidth}
	\vspace{-12pt}
	\centering
	\caption{Mean resolution (\%) across three independently trained students per dataset.}
	\vspace{-0.1in}
	\label{tab:selection-students-main}
	\footnotesize
	\setlength{\tabcolsep}{4pt}
	\begin{tabularx}{\linewidth}{@{}Xrr@{}}
		\toprule
		Dataset & Multilingual & SWE-Pro \\
		\midrule
		SeedBank & 63.33 & 41.22 \\
		Top-5-of-10 & 64.11 & 41.86 \\
		\textsc{SWE-Trail-H5} & 65.67 & 42.82 \\
		\mainrow Top-5-hybrid & \textbf{66.22} & \textbf{43.14} \\
		\bottomrule
	\end{tabularx}
	\vspace{-12pt}
\end{wraptable}
\textbf{Selection alone does not match H5.}
To test whether selection from a larger candidate pool of independent rollouts can match SWE-Trail, Top-5-of-10 samples ten independent trajectories per issue and selects five for training. As shown in Table~\ref{tab:selection-students-main}, it improves mean resolution over SeedBank but remains below H5 by 1.56 points on Multilingual and 0.96 points on SWE-Pro. Thus, selecting from more independent trajectories does not recover the gains from SWE-Trail. The collection process matters beyond post-hoc selection. Detailed results are reported in Table~\ref{tab:selection-students}.

\textbf{Selection and SWE-Trail are complementary.} Top-5-hybrid applies the same verifier to five SeedBank and five H5 trajectories per issue, then selects five trajectories for training. Like Top-5-of-10, it retains 1,500 trajectories in total. The resulting student reaches 66.22 on Multilingual and 43.14 on SWE-Pro, exceeding H5 by 0.56 and 0.32 points, respectively. These results show that post-hoc selection can complement SWE-Trail when applied to the combined candidate set. Appendix~\ref{app:bon-audit} gives the selection rule and candidate counts.

\Needspace{13\baselineskip}
\vspace{-0.05in}
\subsection{What History Adds to Collection}
\label{sec:corpus-results}
\vspace{-0.05in}
\begin{wraptable}{r}{0.46\textwidth}
	\centering
	\captionsetup{width=\linewidth}
	\caption{Collection outcomes over 1,500 trajectories per method.
		Changes in the H5 row are relative to H0.}
		\vspace{-0.1in}
	\label{tab:history-acquisition}
	\label{tab:formal-acquisition-yield}
	\footnotesize
	\setlength{\tabcolsep}{3pt}
	\begin{tabularx}{\linewidth}{@{}Xrrr@{}}
		\toprule
		Method & Resolved & \shortstack{Solved\\issues}
		& \shortstack{Verified\\positive} \\
		\midrule
		SeedBank & 292 & 88 & 263 \\
		\textsc{SWE-Trail-H0} & 329 & 92 & 272 \\
		\mainrow \textbf{\textsc{SWE-Trail-H5}}
		& \textbf{345}\tablegain{+16}
		& \textbf{98}\tablegain{+6}
		& \textbf{285}\tablegain{+13} \\
		\bottomrule
	\end{tabularx}
\end{wraptable}
\textbf{Collection success does not explain student gains.} Table~\ref{tab:history-acquisition} compares the collection outcomes. Relative to SeedBank, H0 produces 37 more resolved trajectories, but improves the student by only 0.11 points on Multilingual. Adding history in H5 produces 16 more resolved trajectories and solves six more issues than H0, while improving the student by a further 2.22 points. Across both benchmarks, the downstream gain is not proportional to the increase in resolved trajectories.

\textbf{History changes the repair process.}
We examine how H5 and H0 trajectories differ beyond their final outcomes using the process statistics in Table~\ref{tab:corpus-process-means} of Appendix~\ref{app:extended-results}.
Relative to H0, H5 reduces the mean number of actions from 40.89 to 39.51 and repeated file reads from 7.59 to 7.01.
The share of edited trajectories with a check after the final edit increases slightly from 88.91\% to 89.53\%.
The clearest shift is toward shorter repair processes with fewer repeated reads, while post-final-edit checking changes only slightly.
Appendix~\ref{app:process-distributions} gives the full distributions.
Similar differences in actions and repeated reads also appear in the students' evaluation trajectories.
On both benchmarks, H5-trained 124B students take fewer actions and repeat fewer file reads on average than H0-trained students.
Appendix~\ref{app:student-eval-behavior} reports the comparison.

\textbf{Most collection gains come from mixed histories.}
Table~\ref{tab:h0-h5-source-regime-yield} in Appendix~\ref{app:extended-results} breaks down the gains in Table~\ref{tab:history-acquisition} by history type.
Stable-solved issues show no difference in resolved trajectories between H0 and H5.
Although mixed histories account for only 51 of the 300 collection issues, they contribute 13 of H5's 16 additional resolved trajectories and five of its six additional solved issues.
Never-solved histories account for the remaining three trajectories and one issue.
The gains are therefore concentrated on issues whose prior attempts include both successes and failures.
This is consistent with Section~\ref{sec:mixed-boundaries}, where such histories reveal decision points that may benefit from additional collection effort.

\Needspace{27\baselineskip}
\subsection{How the Collection Mechanism Changes}
\label{sec:history-mechanism-results}

\begin{wrapfigure}{r}{0.48\textwidth}
	\vspace{-15pt}
	\centering
	\includegraphics[width=\linewidth]{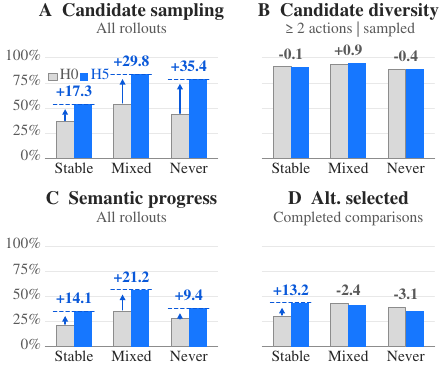}
	\vspace{-0.3in}
\caption{\textbf{History leads to more comparisons.}
	Conditional on candidate sampling, candidate diversity remains similar, as does the alternative-selection rate.
	Table~\ref{tab:branch-funnel-exact} in Appendix~\ref{app:regime-records} gives the full funnel and counts by history type.}
	\label{fig:branch-funnel}
	\vspace{-15pt}
\end{wrapfigure}

\textbf{History leads to more comparisons.} Figure~\ref{fig:branch-funnel} compares H0 and H5 across the stages of collection. With history, candidate sampling occurs in 1,147 of 1,500 rollouts, compared with 664 under H0. Both suffixes are executed in 968 rollouts under H5 and 560 under H0. Panel A shows that candidate sampling increases by 29.8 percentage points on mixed histories and 35.4 on never-solved histories, compared with 17.3 on stable-solved histories. This pattern is consistent with using disagreements and repeated failures to identify where more collection effort may be useful.

\textbf{More comparisons, similar selection rates.} Panel B shows similar candidate diversity under H0 and H5, with at least two distinct actions in about 90\% of candidate-sampling cases. The increase in comparisons therefore comes mainly from candidate sampling occurring more often. Panel C shows better local progress on one route in 40.7\% of H5 rollouts and 28.7\% of H0 rollouts. Yet Panel D shows similar alternative-selection rates among completed comparisons, at 37.4\% and 39.0\%, respectively. Together, Panels C and D show that H5 more often reaches comparisons with clear progress, while the rate of selecting the alternative remains similar. 

%% file: sections/07_related_work.tex
\section{Related Work}
\label{sec:related}
Three lines of work relate to SWE-Trail. Appendix~\ref{app:related} gives the extended discussion.

\textbf{Test-time search and selection.}
Search methods spend extra compute to solve the current issue.
Some sample several independent rollouts and select among them using tests or model judgments~\citep{ehrlich2025codemonkeys}.
Others reuse earlier prefixes or summaries of prior attempts to guide later rollouts on the same issue~\citep{ding2026swereplay,kim2026scalingagentic}.
Tree search explores alternatives within a growing execution tree during the current solve~\citep{kocsis2006bandit,zhou2023lats,antoniades2024swesearch}.
Instead, SWE-Trail uses prior attempts on the same issue to guide where additional collection effort is spent during trajectory generation.
The coding agent proposes actions from its current context, SWE-Trail briefly compares them from the same checkpoint, and the selected route becomes the training trajectory.

\textbf{Trajectory generation and curation.}
Methods for constructing training data generate more tasks and trajectories or increase trajectory diversity through adaptive synthesis~\citep{yang2025swesmith,chen2026tdscaling}.
Others select resolved trajectories or modify completed trajectories through filtering and process-level editing~\citep{pan2024swegym,tao2026swelego,ma2026patches,han2026swetrace}.
Recent work also evolves curation rules using downstream failures~\citep{wang2026curateevo}.
Instead, SWE-Trail uses prior attempts on the same issue to guide where additional collection effort is spent during trajectory generation.
These comparisons collect new execution evidence rather than only selecting or modifying completed trajectories.
Post-hoc selection can still follow, as Section~\ref{sec:cost-results} shows.

\textbf{Experience-guided agents.}
Agents use reflections, distilled insights, workflows, and skills to guide later actions~\citep{shinn2023reflexion,zhao2023expel,wang2024awm,li2026codeskill}. For coding tasks, this experience can come from earlier repair trajectories, human debugging records, or the ongoing attempt~\citep{chen2025sweexp,wang2026memgovern,hayashi2025sage,lin2025seagent}. These methods use experience to guide the agent's reasoning or actions. Instead, SWE-Trail uses history to guide where additional collection effort is spent.
The coding agent proposes actions from its ordinary context without receiving that history.

%% file: sections/08_discussion.tex
\section{Discussion and Limitations}
\label{sec:discussion}

\textbf{What the results show.}
The experiments separate collection success from downstream training value.
H0 resolves more trajectories than independent sampling but brings little gain
on Multilingual, while adding history in H5 produces larger student gains.
Randomizing either where additional collection effort is spent or which
executed route continues lowers mean resolution relative to H5.
The gains also appear with Qwen3-14B-Base.
These results support using history as a collector-side signal for
improving the training value of new trajectories, rather than as advice shown
to the coding agent.

\textbf{Limitations and future work.} Our experiments use two student models but one collection model and one agent scaffold. Future work could test other collection models and scaffolds. Our current setting also keeps history fixed during each collection round and uses the same collection budget for every issue. Future work could update history across rounds and allocate effort differently across issues. The exploratory replay audit in Appendix~\ref{app:selector-replay} also motivates testing more diverse candidate proposals. Extending SWE-Trail beyond coding would require environments that support state restoration and feedback for comparing alternatives. Web navigation is one possible direction.

%% file: sections/09_conclusion.tex
\section{Conclusion}
\label{sec:conclusion}
Our results point to a broader lesson: prior attempts are useful not only for what they contain, but also for guiding where new collection effort should be spent. SWE-Trail operationalizes this idea by compiling prior attempts into rules that determine when additional collection effort may be useful during a new rollout. The coding agent proposes actions without seeing the history, while the collector briefly executes candidates from the same checkpoint and uses the resulting evidence to choose which route to continue. With the training setup fixed, SWE-Trail improves the 124B student's mean resolution over independent sampling by 2.33 percentage points on SWE-bench Multilingual and 1.60 points on SWE-bench Pro. The gains also appear with Qwen3-14B-Base. The H0 comparison shows that collection success alone does not explain the downstream gains. Randomizing either where additional collection effort is spent or which executed route continues lowers mean resolution relative to H5. Together, these results show that prior attempts can improve the training value of new trajectories by guiding collection without being shown to the coding agent as advice.
\newpage

%% file: appendix/appendix.tex
\section*{Overview of the Appendix}

The appendix follows the paper from motivation and collection design to
student outcomes, collection diagnostics, and cost. Table~\ref{tab:appendix-roadmap}
maps the main-text questions to their supporting sections. 

\startcontents[appendix]
\titlecontents{section}[1.8em]
  {\addvspace{5pt}\bfseries}
  {\contentslabel{1.8em}}
  {\hspace*{-1.8em}}
  {\titlerule*[0.6pc]{.}\contentspage}
\printcontents[appendix]{}{1}{\setcounter{tocdepth}{1}}

\begin{table}[!htbp]
  \centering
  \caption{Where each main-text claim is documented in full.}
  \vspace{0.05in}
  \label{tab:appendix-roadmap}
  \small
  \begin{tabular}{@{}p{0.64\textwidth}p{0.30\textwidth}@{}}
    \toprule
    Main text & Appendix \\
    \midrule
    Section~\ref{sec:marginal-coverage}: marginal coverage of independent rollouts & Appendix~\ref{app:target-coverage} \\
    Section~\ref{sec:mixed-boundaries}: mixed-outcome audit and its replication & Appendices~\ref{app:mixed-route-audit} and~\ref{app:cross-model-route-audit} \\
    Section~\ref{sec:method}: compiler, runtime, selector, export, prompts & Appendix~\ref{app:offline-compiler} \\
    Section~\ref{sec:experiments}: issue pool, conditions, training, evaluation & Appendices~\ref{app:data} and~\ref{app:training} \\
    Section~\ref{sec:student-results}: main runs, paired contrasts, language breakdown & Appendix~\ref{app:downstream-results} \\
    Section~\ref{sec:student-results}: second student & Appendix~\ref{app:second-student} \\
    Sections~\ref{sec:corpus-results} and~\ref{sec:history-mechanism-results}: yield, process, and comparison by history type & Appendix~\ref{app:regime-records} \\
    Sections~\ref{sec:student-results} and~\ref{sec:history-mechanism-results}: randomized component controls and replay audit & Appendix~\ref{app:selector-replay} \\
    Section~\ref{sec:cost-results}: history scale, cost, stricter selector, Best-of-$N$ & Appendices~\ref{app:cost}, \ref{app:selector-replay}, and~\ref{app:bon-audit} \\
    Sections~\ref{sec:introduction}, \ref{sec:evidence-selection}, and~\ref{sec:history-mechanism-results}: running Rust example & Appendix~\ref{app:rescue-case} \\
    Sections~\ref{sec:introduction}, \ref{sec:experiments}, and~\ref{sec:student-results}: agent-visible advice & Appendices~\ref{app:condition-mapping} and~\ref{app:advice-exports} \\
    Section~\ref{sec:related}: related work & Appendix~\ref{app:related} \\
    \bottomrule
  \end{tabular}
\end{table}
\section{Evidence Behind the Motivation}
\label{app:motivation-analyses}

\subsection{Coverage and Overlap of Independent Attempts}
\label{app:target-coverage}

We quantify how issue and patch-target coverage grows as more independent
attempts are included. Figure~\ref{fig:motivation-gap}A reports issue coverage,
and Table~\ref{tab:patch-target-coverage} gives the patch-target coverage shown
in panel B.

For issue $i$ with $s_i$ resolved attempts among five, the expected number of
issues solved by a $k$-attempt subset is
\begin{equation}
	\sum_i \left[1-\frac{\binom{5-s_i}{k}}{\binom{5}{k}}\right],
\end{equation}
with an empty numerator treated as zero. This gives 58.4, 71.5, 79.5, 84.6,
and 88.0 expected solved issues for $k=1,\ldots,5$, as reported in
Figure~\ref{fig:motivation-gap}A.

\begin{table}[!htbp]
	\centering
	\caption{Expected patch-target coverage over exact subsets of the five
		SeedBank attempts. Values are the mean fraction of each issue's five-attempt
		union.}
		  \vspace{0.05in}
	\label{tab:patch-target-coverage}
	\small
	\begin{tabular}{@{}lrrrrr@{}}
		\toprule
		History type & $k=1$ & $k=2$ & $k=3$ & $k=4$ & $k=5$ \\
		\midrule
		Stable-solved & 89.96\% & 93.98\% & 96.58\% & 98.38\% & 100\% \\
		Mixed & 73.20\% & 86.42\% & 92.96\% & 96.98\% & 100\% \\
		Never-solved & 80.09\% & 88.89\% & 93.75\% & 97.19\% & 100\% \\
		\bottomrule
	\end{tabular}
\end{table}

Pairwise patch-target Jaccard averages 0.918 on stable-solved, 0.745 on mixed,
and 0.841 on never-solved issues. Among the 198 never-solved issues with a
defined overlap value, 133 have an overlap of at least 0.8 and 93 have an
overlap of 1.0. The remaining 14 issues have no value because attempts that
produce no patch provide no patch-target set to compare. These results show
substantial target overlap even when all attempts fail. They do not explain
why the attempts fail, but they show that independent sampling often returns
to the same repair targets.

\textbf{Behavior-level redundancy.}
Overlapping repair targets do not imply that attempts follow identical action
sequences. We therefore also measure behavioral similarity using
\texttt{SequenceMatcher} over the action-token sequences of all
$\binom{5}{2}$ attempt pairs per issue, giving 3,000 pairs in total.
Actions are tokenized as in Appendix~\ref{app:mixed-route-audit}.
Pairwise values are averaged within each issue and then across issues by
history type. Table~\ref{tab:behavior-redundancy} reports the results.

Mixed histories have the lowest mean similarity on every measure, followed by
never-solved and stable-solved histories. This ordering matches the patch-target
overlap in Figure~\ref{fig:motivation-gap}B. Excluding the 348 of 1,500
attempts that fail or time out in the collector preserves the ordering and
increases the mixed-to-never-solved $d_{\mathrm{full}}$ gap from 0.012 to
0.026. These comparisons use issue-level means. Mixed and never-solved histories
reverse order under pooled-pair medians and under medians that exclude
timed-out and errored attempts.

On never-solved issues, high patch-target overlap coexists with a full-sequence
similarity of 0.503. Failed attempts often return to the same repair targets
without simply repeating the same action sequence.

\begin{table}[!htbp]
	\centering
	\caption{Behavior-level redundancy of the five SeedBank attempts. Pairwise
		similarity is averaged within each issue and then across issues by history
		type. $d_{\mathrm{full}}$ and $d_{25}$ are \texttt{SequenceMatcher} ratios
		over the full and first-25 action-token sequences. $J_{\mathrm{explore}}$
		and $J_{\mathrm{patch}}$ are Jaccard overlaps of explored and modified
		targets, with $J_{\mathrm{patch}}$ computed over issues with a defined value.}
		  \vspace{0.05in}
	\label{tab:behavior-redundancy}
	\small
	\begin{tabular}{@{}lrrrrrr@{}}
		\toprule
		History type & Issues & Pairs & $d_{\mathrm{full}}$ & $d_{25}$ & $J_{\mathrm{explore}}$ & $J_{\mathrm{patch}}$ \\
		\midrule
		Stable-solved & 37 & 370 & 0.605 & 0.629 & 0.665 & 0.918 \\
		Mixed & 51 & 510 & 0.491 & 0.560 & 0.625 & 0.745 \\
		Never-solved & 212 & 2,120 & 0.503 & 0.571 & 0.642 & 0.841 \\
		\bottomrule
	\end{tabular}
\end{table}

The coverage and overlap results show where additional independent attempts
become increasingly redundant. They do not show whether earlier attempts can
identify where additional collection effort may be useful. We examine that
question next.

\subsection{Decision Differences in Mixed Histories}
\label{app:mixed-route-audit}

Mixed histories are informative because they contain both successful and failed
attempts on the same issue. The audit asks whether these histories reveal
decision points where more than one next action is reasonable given the
available evidence. It uses outcomes first to locate a meaningful difference
between a successful and a failed attempt. It then removes the outcomes and
future feedback and asks whether the two states are comparable and whether both
next actions are defensible from the information available at that point.
This directly tests the allocation signal studied in
Section~\ref{sec:mixed-boundaries}. The issue set, pair construction, model
inputs, label spaces, and aggregation rule were fixed before the results were
computed. The audit model was \texttt{GLM-5.1} at temperature zero.

\textbf{Pair construction.}
The input contains the 51 mixed issues whose five SeedBank attempts include at
least one resolved and one unresolved trajectory. Here, \emph{native} means
generated by the coding agent from its ordinary view, with no collector text.
Canonical native events are converted into 1-based action timelines after
planning-only events are removed. Each action records its category, repository
target, normalized command or edit, the immediately preceding native reasoning,
and the resulting observation and error flag. Whitespace is normalized.
Problem statements are capped at 7,000 characters, per-action native reasoning
at 900, observations at 800, edit summaries at 1,700, and other action
summaries at 700.

For every issue, we score every resolved/unresolved pair. When a naturally
completed unresolved trajectory exists, collector failures and timeouts are
excluded from the candidate pool. Otherwise, the censored unresolved attempts
form the pool. This produces 43 primary pairs and eight explicitly censored
pairs.

The audit compares a successful and failed attempt that reached similar
observed situations. We select the highest-scoring pair for each issue.
Let $d_{\mathrm{full}}$ be the \texttt{SequenceMatcher} ratio between the full
action-token sequences and $d_{25}$ the ratio between their first 25 actions.
An action token is its action category and normalized target. Terminal,
environment, and validation actions additionally include the recognized command
family. Let $J_{\mathrm{explore}}$ and $J_{\mathrm{patch}}$ be Jaccard overlap
of explored and modified repository targets. The score is
\begin{equation}
	s = 0.45d_{\mathrm{full}} + 0.30d_{25}
	+ 0.15J_{\mathrm{explore}} + 0.10J_{\mathrm{patch}}.
	\label{eq:mixed-audit-pair-score}
\end{equation}
The maximum-scoring pair is selected, with deterministic tie-breaking by fewer
coding actions and then run ID.

\textbf{Stage 1: retrospective localization.}
The first stage uses the complete trajectories and their outcomes only to
locate an early decision difference that may matter for the repair route.
The model receives the problem statement, both complete action timelines,
their outcomes, the censor flag, and the pair score. It returns a status,
one of \texttt{found}, \texttt{no\_meaningful\_divergence}, and
\texttt{not\_comparable}, together with a rationale and a confidence.
For a found case, it also returns a boundary index for each route, summaries
of the shared state and the two route commitments, a divergence type, and an
outcome-relevance label. That label is one of \texttt{direct},
\texttt{plausible}, \texttt{weak}, and \texttt{none}. Returned indices are
rejected if they fall outside the supplied timelines. The audit uses the
following exact system instruction.

\begin{sweprompt}{Stage 1: retrospective-localization instruction}
	Retrospectively compare one resolved and one unresolved native coding-agent
	trajectory for the same issue. Locate the earliest outcome-relevant persistent
	route divergence. Ignore task-tracker updates, harmless ordering differences,
	equivalent reads, and wording changes. A meaningful divergence changes the
	working hypothesis, repository scope, edit semantics, validation strategy,
	failure recovery, or stopping decision. Outcomes are provided only to locate
	the boundary; do not infer a gold patch or claim that every difference caused
	the outcome. If the trajectories never reach comparable states or no meaningful
	divergence can be supported, say so. Cite only supplied action indices. Return
	one JSON object and no prose outside it.
\end{sweprompt}

The full divergence-type label space is localization, hypothesis,
implementation, validation, recovery, finish, resource/infra, and other.
Only hypothesis, implementation, and validation occur among the 44 localized
cases.

\textbf{Stage 2: outcome-blind action audit.}
Only a Stage-1 \texttt{found} case proceeds to Stage 2. For each source route,
the model sees the prefix strictly before its localized boundary, including
the feedback already available at that point. It also sees the immediate
candidate action with its observation removed. It does not receive either
trajectory outcome, candidate feedback, future actions, the gold patch, the
Stage-1 retrospective rationale, or the solved/failed source identity.
The two sources are exposed only as routes A and B, with their source
identities hidden.

This stage asks whether the visible states are sufficiently comparable and
whether both immediate actions are reasonable using only the information
available at those states. The audit uses the following exact system
instruction.

\begin{sweprompt}{Stage 2: prospective branch-audit instruction}
	Audit whether two native coding decisions represented a real prospective
	fork state. The two prefixes come from independent attempts on the same issue,
	and route labels are opaque. You do not know either outcome or any candidate
	feedback. First decide whether the visible states are sufficiently comparable.
	Then judge each immediate candidate only from its own visible prefix: is the
	action supported, and could it obtain decision-relevant information? Two choices
	are reasonable only when both are prospectively defensible, semantically
	distinct, and neither depends on hidden future evidence. Do not reward mere
	activity, extra length, or knowledge of a gold solution. Return one JSON object
	and no prose outside it.
\end{sweprompt}

The structured response records five labels, a reason, and a confidence.
State comparability is equivalent, similar with minor differences, or different.
Route distinctness is distinct, equivalent, or unclear. Support for each
candidate is supported, partially supported, or unsupported. Information gain
for each candidate is high, medium, low, or none. The schema calls the final
field prospective branchability. Its labels are two reasonable routes,
one reasonable route, equivalent routes, non-comparable, and insufficient
evidence. The call is then repeated with A/B order reversed.

\textbf{Execution, aggregation, and missingness.}
We aggregate the two route-order judgments before counting supported decision
points. Two \texttt{two\_reasonable\_routes} votes produce the strict label.
One such vote produces the possible-two-routes label. Two one-route votes
produce a one-route consensus only when they retain the same source-level
support pattern. Otherwise, the result is order-sensitive. Other identical
votes retain their shared branchability label, and any remaining disagreement
is order-sensitive.

Stage 1 returns valid structured results for 47/51 pairs. These comprise
44 localized divergences and three no-meaningful-divergence cases.
The remaining four are collector-censored pairs whose audit calls failed after
retries. None is imputed as a negative. Both orderings complete for all
44 localized cases, giving 88 valid Stage-2 responses. In the primary set,
this gives 43 complete dispositions, 40 localized cases, and three
no-meaningful-divergence cases. The eight censored pairs are tabulated
separately.

In the primary set, 21 strict and ten possible cases give 31 decision points
with evidence that more than one next action may be worth comparing.
Order-sensitive cases do not enter this count.
Table~\ref{tab:mixed-audit-funnel} gives the complete funnel.

\begin{table}[!htbp]
	\centering
	\caption{Complete primary audit funnel. ``Share'' retains the original
		51-issue denominator, while ``Retention'' uses the preceding row. The eight
		collector-censored pairs remain in
		Table~\ref{tab:mixed-audit-classification-counts} but do not enter the
		primary funnel after pair selection.}
		  \vspace{0.05in}
	\label{tab:mixed-audit-funnel}
	\small
	\begin{tabular}{@{}lrrr@{}}
		\toprule
		Stage & Count & Share of 51 & Retention \\
		\midrule
		Selected mixed-issue pairs & 51 & 100.0\% & 100.0\% \\
		Naturally completed unresolved pair & 43 & 84.3\% & 84.3\% \\
		Complete audit disposition & 43 & 84.3\% & 100.0\% \\
		Localized divergence and dual-order audit & 40 & 78.4\% & 93.0\% \\
		\mainrow Decision point supporting comparison & 31 & 60.8\% & 77.5\% \\
		\bottomrule
	\end{tabular}
\end{table}

\begin{table}[!htbp]
	\centering
	\caption{Complete aggregate audit classifications. Divergence type and
		outcome relevance partition the Stage-1 \texttt{found} rows, while final
		disposition partitions every pair. ``Primary'' uses naturally completed
		unresolved attempts, and ``censored'' contains collector failures or
		timeouts.}
		  \vspace{0.05in}
	\label{tab:mixed-audit-classification-counts}
	\small
	\begin{tabular}{@{}llrrr@{}}
		\toprule
		Family & Classification & All (51) & Primary (43) & Censored (8) \\
		\midrule
		\multirow{3}{*}{Stage 1}
		& Localized divergence & 44 & 40 & 4 \\
		& No meaningful divergence & 3 & 3 & 0 \\
		& Not analyzed: API long tail & 4 & 0 & 4 \\
		\midrule
		\multirow{3}{*}{Divergence type}
		& Hypothesis & 3 & 3 & 0 \\
		& Implementation & 28 & 25 & 3 \\
		& Validation & 13 & 12 & 1 \\
		\midrule
		\multirow{3}{*}{Outcome relevance}
		& Direct & 25 & 21 & 4 \\
		& Plausible & 17 & 17 & 0 \\
		& Weak & 2 & 2 & 0 \\
		\midrule
		\multirow{7}{*}{Final disposition}
		& Strict two reasonable routes & 23 & 21 & 2 \\
		& Possible two reasonable routes & 11 & 10 & 1 \\
		& One reasonable route, consensus & 3 & 2 & 1 \\
		& Order-sensitive & 6 & 6 & 0 \\
		& Equivalent routes & 1 & 1 & 0 \\
		& No meaningful divergence & 3 & 3 & 0 \\
		& Not analyzed: API long tail & 4 & 0 & 4 \\
		\bottomrule
	\end{tabular}
\end{table}

\subsection{Independent Cross-Model Audit Replication}
\label{app:cross-model-route-audit}

We next test whether the mixed-history audit depends on the model used to
perform it. An independent Auditor B, \texttt{gpt-5.6-sol}~\citep{openai2026gpt56},
replays the same 51-pair set used in
Appendix~\ref{app:mixed-route-audit}. Auditor B never receives Auditor A's
responses, labels, aggregate counts, or paper claims. Its Stage-1 run uses the
same retrospective localization instruction and structured schema. Its Stage-2
run again hides outcomes, candidate feedback, future actions, the gold patch,
and source identity, and evaluates both route orders to control for the
positional bias of generative judges~\citep{sun2026judging}. Calls use medium reasoning
effort and provider-default temperature.

The end-to-end replication independently completes all 43 primary pairs.
Of these, 41 contain a localized divergence and two contain no meaningful
divergence. Auditor B identifies 26 decision points supporting comparison,
comprising 16 strict and ten possible cases. Auditor A identifies 31,
comprising 21 strict and ten possible cases. These aggregate counts do not by
themselves establish agreement on individual cases.

The two auditors agree on \texttt{found} versus
\texttt{no\_meaningful\_divergence} in 40/43 primary and 4/4 commonly completed
censored cases. They both localize 43 cases overall. The solved-route and
failed-route boundary distances each have median zero. Twenty boundary pairs
match exactly, and 26 divergence-type labels match exactly.

To separate disagreement about where the decision difference occurs from
disagreement about the actions themselves, we also compare the auditors at
fixed boundaries. This comparison uses the boundaries from the first 32 pairs
completed by Auditor A, including 28 primary and four censored pairs.
Auditor B judges the same outcome-blind inputs in both route orders. On the
28 primary pairs, Auditor B identifies 16 decision points supporting
comparison, and both auditors support 15.

Table~\ref{tab:cross-model-agreement} compares the final judgments on whether
both actions are defensible. Binary-label agreement is 60.5\% on the
end-to-end primary set and 64.3\% under fixed boundaries. The auditors jointly
support 20 primary cases end to end.

\begin{table}[!htbp]
	\centering
	\caption{Independent cross-model replication of the mixed-history audit.
		$N$ is the common complete-disposition set for each row. Positive means
		strict or possible two reasonable routes. ``Both'' is the conservative
		positive intersection. Pos./Neg. are positive/negative agreement.
		Both auditors complete all 43 primary pairs end to end. Auditor A identifies
		31 supported comparison points and Auditor B identifies 26. Raw, Pos., Neg.,
		$\kappa$, and AC1 describe binary audit labels, not agreement on divergence
		localization or the full label vocabulary.}
		  \vspace{0.05in}
	\label{tab:cross-model-agreement}
	\small
	\resizebox{\textwidth}{!}{%
		\begin{tabular}{@{}llrrrrrrrrr@{}}
			\toprule
			Track & Set & $N$ & A$+$ & B$+$ & Both$+$ & Raw & Pos. & Neg. & $\kappa$ & AC1 \\
			\midrule
			End to end & Primary & 43 & 31 & 26 & 20 & 60.5\% & 70.2\% & 41.4\% & .129 & .285 \\
			End to end & Censored & 4 & 3 & 3 & 3 & 100.0\% & 100.0\% & 100.0\% & 1.000 & 1.000 \\
			Fixed boundary & Primary & 28 & 24 & 16 & 15 & 64.3\% & 75.0\% & 37.5\% & .205 & .397 \\
			Fixed boundary & Censored & 4 & 3 & 2 & 2 & 75.0\% & 80.0\% & 66.7\% & .500 & .529 \\
			\bottomrule
	\end{tabular}}
\end{table}

The audits identify candidate locations for comparison rather than prescribing
which action to take. Agreement is stronger on locating meaningful differences
than on judging whether both actions are reasonable. The compared prefixes
come from independent attempts. During collection, alternatives are instead
executed from the same checkpoint. Appendix~\ref{app:branch-funnel} reports
whether the coding agent proposes distinct actions and whether short execution
provides evidence for choosing between them.

\section{Collector Algorithm, Schemas, and Prompts}
\label{app:offline-compiler}
\label{app:schema}

This appendix gives the implementation details behind Section~\ref{sec:method}.
The collector uses history only to decide where additional collection effort
may be useful and what evidence should distinguish candidate routes.
The coding agent receives its ordinary context and generates all candidate
actions itself.

\subsection{End-to-End Collection Algorithm}
\label{app:algorithm}

\begin{algorithm}[ht]
	\caption{SWE-Trail collection for one issue.
		\algchip{algprivate}{\strut Orange} marks collector-private operations;
		\algchip{algagent}{\strut blue} marks coding-agent generation from its
		ordinary visible context.}
	\label{alg:swe-trails-full}
	\small
	\begin{algorithmic}[1]
		\Require issue $x_i$, ordered history $H_i$, coding agent $\pi$, candidate
		count $K$, suffix horizon $L$
		\State \algrow{algprivate}{$E_i \gets
			\Call{AnalyzeHistory}{x_i,H_i}$}
		\State \algrow{algprivate}{$P_i \gets
			\Call{RunReadOnlyChecks}{x_i,E_i}$}
		\State \algrow{algprivate}{$(R_i,D_i,u_i,\lambda_i) \gets
			\Call{CompileAndFreeze}{x_i,E_i,P_i}$}
		\State initialize agent-visible trajectory $\tau$, workspace $w$, and
		$\mathit{compared}\gets\mathrm{false}$
		\While{$\tau$ is not terminal}
		\State \algrow{algagent}{$(a,o,w) \gets
			\Call{AgentStep}{\pi,\tau,w}$}
		\State append generated action $a$ and observation $o$ to $\tau$
		\State $d \gets \Call{MatchDetector}{D_i,\tau,w}$
		\If{$\neg\mathit{compared}$ and $d\neq\varnothing$}
		\State $c \gets \Call{Checkpoint}{\tau,w}$
		\State \algrow{algagent}{$C \gets
			\Call{SampleCandidates}{\pi,\tau,K,\algchip{algprivate}{$\lambda_i$}}$
			\Comment{$\lambda_i$ sets temperatures}}
		\State $C' \gets \Call{SemanticDeduplicate}{C}$
		\State $a_0 \gets$ default candidate in $C'$
		\State \algrow{algprivate}{$a_1 \gets
			\Call{PrivateShortlist}{C',R_i,d}$}
		\If{$a_1$ exists}
		\State \algrow{algagent}{$(s_0,w_0) \gets
			\Call{ExecuteSuffix}{c,a_0,L}$}
		\State \algrow{algagent}{$(s_1,w_1) \gets
			\Call{ExecuteSuffix}{c,a_1,L}$}
		\State \algrow{algprivate}{$j \gets
			\Call{BlindSelect}{u_i,d,R_i,s_0,s_1}$
			\Comment{$R_i$ is background evidence only}}
		\If{$j$ is tie or invalid}
		\State $j\gets0$
		\EndIf
		\State $(\tau,w) \gets \Call{ContinueFrom}{c,s_j,w_j}$
		\EndIf
		\State $\mathit{compared}\gets\mathrm{true}$
		\EndIf
		\EndWhile
		\State \Return $\Call{ExportContinuousTrajectory}{\tau}$
	\end{algorithmic}
\end{algorithm}

Algorithm~\ref{alg:swe-trails-full} shows the main collection path.
For each detector state $d$, $u_i(d)$ specifies what evidence should be
considered when comparing the executed routes. Policy calls, including
candidate generation and suffix execution, receive only the ordinary issue,
conversation, and repository observations. The schedule $\lambda_i$ changes
sampling temperatures without adding history to the coding-agent prompt.
Shortlisting and selection choose among actions already proposed by the coding
agent. They do not write new actions or replace the agent's reasoning.
Proposal-only rechecking and runtime failures are described in
Appendix~\ref{app:online-runtime}.

\subsection{Collection Cost Accounting}
\label{app:cost-boundary}

We separate the size of the exported training corpus from the compute used to
collect it. Given an existing set of prior attempts, we use two accounting
boundaries:
\begin{equation}
	\begin{aligned}
		C_{\mathrm{overhead}}
		&= C_{\mathrm{compile}} + C_{\mathrm{candidates}}
		+ C_{\mathrm{extra\ suffix}} \\
		&\quad + C_{\mathrm{selection}} + C_{\mathrm{retry}}, \\
		C_{\mathrm{bank}} &= C_{\mathrm{rollout}} + C_{\mathrm{overhead}}.
	\end{aligned}
\end{equation}
Here $C_{\mathrm{overhead}}$ is the additional compute beyond an ordinary
rollout, while $C_{\mathrm{bank}}$ includes both the ordinary rollout and this
overhead. We report model tokens and environment actions under these
boundaries. Summed trajectory-hours are reported separately from concurrent
wall-clock time. Appendix~\ref{app:cost} gives the measured breakdown.
Provider-token accounting is complete for H5 and H10. H0 records runtime calls
but not offline compilation, and SeedBank has no source-provider usage record.
We do not impute unavailable components.

\subsection{Offline Compiler}
\label{app:compiler}
The compiler turns information from earlier attempts into fixed rules for
deciding where additional collection effort may be useful in a new rollout.
It follows the same sequence for each issue. All model-produced fields pass
schema and content checks before they can affect collection.

\begin{swespecpanel}[breakable]
	\begin{enumerate}[
		label=\textcolor{antprimarydark}{\bfseries\arabic*.},
		leftmargin=1.65em,
		labelsep=0.55em,
		itemsep=3pt,
		topsep=0pt,
		parsep=0pt,
		partopsep=0pt
		]
		\item \textcolor{antprimarydark}{\textbf{History.}}
		Read the ordered source attempts and their observed outcomes.
		
		\item \textcolor{antprimarydark}{\textbf{Evidence.}}
		Extract supported patterns from the history, including decisions repeated
		across successful attempts, differences between repair routes, recurring
		failures, exhausted searches, and unresolved questions. Each item cites the
		source events that support it.
		
		\item \textcolor{antprimarydark}{\textbf{Read-only checks.}}
		Run bounded repository checks when the history leaves an unresolved question.
		Each check records its command, observation, interpretation, and provenance.
		Checks cannot edit source code, access a reference patch, or directly create
		a detector rule.
		
		\item \textcolor{antprimarydark}{\textbf{Rules.}}
		Combine the history evidence and read-only checks into candidate detector
		rules. Each rule includes observable conditions, a local comparison objective,
		and explicit abstention conditions.
		
		\item \textcolor{antprimarydark}{\textbf{Admission.}}
		Parse, lint, deduplicate, and replay the candidate rules on the source
		trajectories. Fix all admitted rules before the main collection begins.
	\end{enumerate}
\end{swespecpanel}

Internally, each evidence item records an identifier, source type, hypothesis
or constraint, direct observation, bounded interpretation, uncertainty,
history type, and links to supporting or contradicting evidence.
The accompanying human-readable representation forms the private acquisition
record $R_i$. Neither representation enters a coding-agent request.

\subsection{Detector Schema and Admission Checks}
\label{app:detector-schema}
A detector rule can guide collection only after its rule and supporting
evidence pass the checks below. An \emph{active detector rule} has passed
these checks and is fixed for collection. At runtime, the rule is evaluated
against the actions and observations seen so far.

Each detector rule contains the following fields.
\begin{swespecpanel}[breakable]
	\begin{description}[
		style=multiline,
		leftmargin=0.23\linewidth,
		labelwidth=0.18\linewidth,
		labelsep=0.05\linewidth,
		font=\normalfont\bfseries\color{antprimarydark},
		itemsep=2pt,
		topsep=0pt,
		parsep=0pt,
		partopsep=0pt
		]
		\item[Scope] A state and decision family.
		\item[Predicates] Observable conditions with a minimum-observation window.
		\item[Support] Evidence IDs and the rationale for spending additional
		collection effort at this state.
		\item[Comparison] A local objective for comparing executed routes.
		\item[Status] Confidence, replay coverage, and admission status.
		\item[Fallback] Explicit abstention and fallback conditions.
	\end{description}
\end{swespecpanel}

Executable predicates may use only features available during the current
rollout. They cannot contain a reference patch, hidden tests, historical
solution actions, repository-specific instructions to the coding agent, or
future outcome labels. H5 and H10 rules must cite history evidence and pass
history replay. H0 rules must cite source-clean repository evidence.
Parse failure, prohibited content, contradictory rationale, duplicate
signatures, or insufficient support causes abstention.

H5 compiles 687 candidate detector rules. Of these, 569 match at least one
source trajectory during replay, and 397 pass the admission checks for
collection. Appendix~\ref{app:rescue-case} gives the concrete post-edit
read-without-validation rule used in the running Rust example.

\subsection{Online Runtime Details}
\label{app:online-runtime}

The runtime evaluates the detector after each new repository observation.
A trajectory executes at most one comparison. If all sampled candidates are
equivalent, no comparison is executed. The collector may sample candidates
once more after the default action runs.

\begin{swespecpanel}[breakable]
	\begin{enumerate}[
		label=\textcolor{antprimarydark}{\bfseries\arabic*.},
		leftmargin=1.65em,
		labelsep=0.55em,
		itemsep=3pt,
		topsep=0pt,
		parsep=0pt,
		partopsep=0pt
		]
		\item \textcolor{antprimarydark}{\textbf{Checkpoint.}}
		Snapshot the agent-visible conversation, task workspace, and runtime metadata
		when a detector rule matches the current rollout.
		
		\item \textcolor{antprimarydark}{\textbf{Candidates.}}
		Sample eight one-step continuations from the same model weights and visible
		prefix. Candidate zero uses temperature 0.7. H0 uses 0.7 for every slot.
		Within H5 and H10, stable-solved histories use 0.7 throughout, mixed
		histories alternate 0.7/0.85, and never-solved histories cycle
		0.7/0.85/1.0.
		
		\item \textcolor{antprimarydark}{\textbf{Shortlist.}}
		Cluster semantically equivalent proposals. Keep candidate zero and privately
		shortlist at most one distinct, executable alternative.
		
		\item \textcolor{antprimarydark}{\textbf{Execution.}}
		Restore two isolated workspaces from the same checkpoint and execute the
		default and alternative candidates for two observations. If restoration or
		execution fails, retain candidate zero.
		
		\item \textcolor{antprimarydark}{\textbf{Selection.}}
		Shuffle the two route identities and compare their reasoning, actions,
		workspace changes, and repository observations using the local comparison
		objective. The private record provides background evidence only. Select a
		route only when the executed suffixes provide clear evidence that one makes
		better progress. Otherwise, retain the default route.
		
		\item \textcolor{antprimarydark}{\textbf{Export.}}
		Continue only from the selected workspace and export one continuous
		trajectory. Preserve the unselected suffix only in the private collection
		record.
	\end{enumerate}
\end{swespecpanel}

\subsection{Information Boundaries and Export Checks}
\label{app:contracts}
\label{app:export}
The exported trajectory must preserve the coding agent's original reasoning
and actions without adding private collector information.
Table~\ref{tab:visibility-contract} summarizes which information each stage
may use. The export checks enforce this boundary and verify that the retained
actions and tool results form one continuous trajectory.
\begin{table}[!htbp]
	\centering
	\caption{Information boundary enforced during collection and export.
		``Private'' means that information may inform the collector but never enters
		a coding-agent request. The last column contains the two executed suffixes
		used by the route selector.}
		\vspace{0.051in}
	\label{tab:visibility-contract}
	\small
	\resizebox{\textwidth}{!}{%
		\begin{tabular}{@{}lcccc@{}}
			\toprule
			Component & Visible prefix & Prior attempts & Private record & Executed suffix pair \\
			\midrule
			Offline compiler & issue only & H-dependent & yes & no \\
			Candidate generation & yes & no & no & no \\
			Private shortlist & yes & yes & yes & no \\
			Blind suffix selector & yes & private & private & yes \\
			Exported trajectory & selected route & no & no & selected only \\
			\bottomrule
	\end{tabular}}
\end{table}

The private comparison log records the visible prefix, shared checkpoint,
candidate actions, both executed suffixes, and the selector's decision.
Isolation checks require identical prefix and pre-execution workspace digests
and separate workspaces after restoration. A failed check prevents the
alternative route from being selected.

Export canonicalization restores the coding agent's original reasoning attached
to each selected action, pairs every tool call with exactly one result, and
removes private collector events. It does not rewrite or reconstruct reasoning.
A record is rejected if a selected event cannot be traced to the raw stream,
tool pairing fails, the exported patch differs from the selected workspace,
or private collector content appears in the training trajectory.

\subsection{Prompts and Inputs}
\label{app:prompts}

Table~\ref{tab:prompt-interfaces} gives the input and output boundary for each
model-assisted collector stage. Candidate generation is intentionally absent.
It uses the ordinary coding-agent prompt and visible trajectory prefix, with
no SWE-Trail instruction and no history. The system instructions below are
reproduced exactly as used during collection.

\begingroup
	\setlength{\tabcolsep}{3.5pt}
	\setlength{\LTleft}{0pt plus 1fill}
	\setlength{\LTright}{0pt plus 1fill}
	\setlength{\LTcapwidth}{\textwidth}
	\setlength{\LTpre}{\intextsep}
	\setlength{\LTpost}{\intextsep}
	\renewcommand{\arraystretch}{1.1}
	\begin{longtable}{@{}
		>{\small\raggedright\arraybackslash}p{\dimexpr.195\textwidth-1.17\tabcolsep\relax}
		>{\small\raggedright\arraybackslash}p{\dimexpr.2825\textwidth-1.695\tabcolsep\relax}
		>{\small\raggedright\arraybackslash}p{\dimexpr.3175\textwidth-1.905\tabcolsep\relax}
		>{\small\raggedright\arraybackslash}p{\dimexpr.205\textwidth-1.23\tabcolsep\relax}@{}}
		\caption{Inputs and outputs of the model-assisted collector stages.
			All outputs remain private to the collector and are never inserted into a
			coding-agent request.}
		\label{tab:prompt-interfaces}\\
		\toprule
		Stage & Structured inputs & Required output & Downstream consumer \\
		\midrule
		\endfirsthead
		\multicolumn{4}{@{}l}{\small Table~\ref{tab:prompt-interfaces} (continued)}\\
		\toprule
		Stage & Structured inputs & Required output & Downstream consumer \\
		\midrule
		\endhead
		\midrule
		\endfoot
		\bottomrule
		\endlastfoot
			History diagnosis
			& issue, ordered attempts, outcomes
			& mechanisms, decision differences, unresolved questions
			& repository-check planner \\
			\addlinespace[2pt]
			Repository exploration
			& diagnosis, repository snapshot
			& bounded commands, findings, provenance
			& evidence auditor \\
			\addlinespace[2pt]
			Evidence audit
			& diagnosis, repository observations
			& supported/contradicted claims, uncertainty
			& rule synthesis \\
			\addlinespace[2pt]
			State synthesis/compiler
			& audited evidence, feature schema
			& states, predicates, objectives, abstentions
			& private detector \\
			\addlinespace[2pt]
			Candidate shortlist
			& visible prefix, candidates, private record
			& default plus one existing candidate index
			& matched executor \\
			\addlinespace[2pt]
			Suffix selector
			& shuffled suffixes, comparison objective, private record as prior
			& select A, select B, or tie with evidence
			& trajectory continuation \\
	\end{longtable}
\endgroup

\begin{sweprompt}{History diagnosis: system instruction}
	Audit same-issue coding trajectories before any environment exploration.
	Separate successful mechanisms, failed mechanisms, decision boundaries, and
	unresolved repository questions. Failed trajectories are counterexamples. Do
	not generate detector code, advice, file hints, or a patch. Return JSON only.
\end{sweprompt}

\begin{sweprompt}{Bounded repository exploration: system instruction}
	Act as a bounded source-clean repository explorer. Start from the supplied
	history diagnosis, refine competing hypotheses, and propose safe read-only or
	validation probes. Do not edit source, use network/history, or generate a
	detector. One command per probe, with no pipes or redirection. Return JSON only.
\end{sweprompt}

\begin{sweprompt}{Evidence audit: system instruction}
	Act as an auditor separate from history diagnosis and repository exploration.
	Admit only claims with exact supplied evidence. Separate observation, supported
	inference, negative boundary, and uncertainty. Do not generate detector code,
	patch text, or a gold answer. Return JSON only.
\end{sweprompt}

\begin{sweprompt}{Acquisition-state synthesis: system instruction}
	Synthesize prospective issue-specific acquisition states only after reviewing
	history diagnosis and audited repository probes. Conditions may use only the supplied public feature
	schema and must not encode repository targets. Explain every condition and cite supplied history and audited probe evidence. Return JSON only.
\end{sweprompt}

\begin{sweprompt}{History-conditioned detector compiler: system instruction}
	Compile an issue-private prospective branch detector from atomic same-issue
	history experiences and, when supplied, audited repository probe experiences.
	History failures are counterexamples. Probe facts refine issue knowledge but
	must not automatically become detector states. Preserve the strong native prior
	for stable-solved issues. Use only declared process features, cite exact
	evidence, and abstain when branch timing is unsupported. Return JSON only.
\end{sweprompt}

For H0, the compiler instead receives the following history-free instruction.

\begin{sweprompt}{H0 detector compiler: system instruction}
	Compile an issue-private prospective branch detector using only the issue
	statement and source-clean repository probe evidence. No historical trajectory,
	outcome regime, or prior detector is available. Use only declared process
	features, cite exact supported probe experiences, and abstain when static
	repository evidence cannot justify a prospective resampling point. Do not
	encode repository paths or patch answers. Return JSON only.
\end{sweprompt}

\begin{sweprompt}{Candidate shortlist: system instruction and user contract}
	Shortlist native coding-agent actions generated from one identical prefix.
	Private history and audited probe evidence may only distinguish targets already
	proposed by a candidate. Candidate 0 is the native control and must remain in
	the matched pair. Do not invent a new action. Return JSON only.
	
	\textnormal{\bfseries User contract.} Return \texttt{provisional\_winner\_index},
	\texttt{ranked\_candidate\_indices}, \texttt{reason}, \texttt{confidence}, and
	\texttt{candidate\_assessments}. Assess visible support, reasoning/action
	alignment, expected information gain, historical redundancy, and tool validity.
	Indices use \texttt{candidate\_index} from \texttt{candidate\_actions}.
\end{sweprompt}

\begin{sweprompt}{Executed-suffix selector: system instruction and user contract}
	Compare two coding-agent suffixes executed from the same conversation and
	workspace checkpoint. Branch A and Branch B are opaque shuffled labels with
	equal status; do not infer which lineage produced either branch. Judge only
	local observed evidence. Private history is a failure/success prior, never a
	gold answer. Select a branch only when its executed suffix demonstrates better
	semantic progress; otherwise return tie. The decision field must be exactly
	\texttt{select\_a}, \texttt{select\_b}, or \texttt{tie}.
	\texttt{semantic\_progress} must be a JSON boolean. Return JSON only.
	
	\textnormal{\bfseries User contract.} Return \texttt{decision}, \texttt{reason},
	\texttt{confidence}, \texttt{semantic\_progress}, and
	\texttt{branch\_assessments}. Each assessment includes \texttt{branch},
	\texttt{new\_evidence}, \texttt{validation\_signal}, \texttt{tool\_error},
	\texttt{unchanged\_retry}, \texttt{unsupported\_finish}, and \texttt{reason}.
\end{sweprompt}

\section{Data and Experimental Conditions}
\label{app:data}

\subsection{Collection Issues and Prior Attempts}
\label{app:collection-pool}

The main experiments use 300 repository-level issues from
SWE-rebench V2~\citep{badertdinov2026swerebenchv2}, with one issue per
repository. Each collection condition produces five trajectories per issue.
SeedBank provides the five independently sampled attempts used as history by
H5. H10 adds five more eligible attempts to the same five, giving ten prior
attempts per issue.

We define three history types from the outcomes of the five SeedBank attempts
before the main collection begins. The stable-solved group contains 37 issues
for which all five attempts succeed. The mixed group contains 51 issues with
both successful and failed attempts. The never-solved group contains 212 issues
for which all five attempts fail. These labels describe the observed SeedBank
outcomes and are separate from how the compiler interprets each history.
Appendix~\ref{app:compiler-records} reports the compiler classes.

Collection runs in an offline sandbox with no network access and no repository
commit history. The coding agent sees the working tree at the issue commit but
cannot access later commits. The reference patch, fixing commit, and externally
published solutions are also unavailable. The read-only repository checks in
Section~\ref{sec:branch-opportunity} follow the same restrictions. Any
information already encoded in the model weights is shared across conditions,
since all collection conditions use the same model on the same issue set.

\subsection{Evaluation Benchmarks and Overlap Audit}
\label{app:overlap-audit}

We evaluate on SWE-bench Multilingual and SWE-bench Pro.
SWE-bench Multilingual, released with SWE-smith~\citep{yang2025swesmith},
contains 300 issues across nine languages: C, C++, Go, Java, JavaScript,
TypeScript, PHP, Ruby, and Rust.
Appendix~\ref{app:per-language} reports results by language.
SWE-bench Pro~\citep{deng2025swebenchpro} contains longer-horizon
repository issues whose reference patches typically span multiple files.
We use its public split of 731 tasks. Both benchmarks use the evaluation
protocol in Appendix~\ref{app:evaluation}.

We verify that the collection and evaluation sets do not share issues or
repositories. Exact instance IDs and canonical lowercase owner/repository
names are compared across the sets. The 300 Multilingual tasks span
41 repositories, and the 731 SWE-Pro tasks span 437 repositories. Neither
benchmark shares an issue or repository with the collection pool.

\subsection{Experimental Conditions}
\label{app:condition-mapping}

The conditions separate three ways of using existing trajectories.
The three curation conditions are controlled adaptations rather than full
reproductions of the original methods. They modify the same completed SeedBank
trajectories without collecting new interactions. SWE-Exp uses prior attempts
to create information that is shown to the coding agent during a new rollout.
SWE-Trail instead keeps prior attempts private to the collector and uses them
to guide where additional collection effort is spent.

Our SWE-Gym adaptation~\citep{pan2024swegym} conditions each record on its
outcome. Successful records receive the outcome annotation while preserving
their original messages and actions. Unresolved records remain unchanged.
The adaptation does not apply a condition-specific loss mask.

Our SWE-Lego adaptation~\citep{tao2026swelego} applies deterministic
tool-error masking to successful records. It masks only assistant turns whose
tool result indicates strict tool misuse, including invalid arguments, missing
paths, permission errors, or shell syntax errors. Failed tests and bug
reproduction output remain trainable. Unresolved records remain unchanged.

Our SWE-TRACE adaptation~\citep{han2026swetrace} removes noncritical messages
from successful records. It retains interactions involving patch-modified files,
validation, final repository inspection, and the final answer. Repeated
non-mutating observations are deduplicated. Messages are removed rather than
loss-masked. Unresolved records remain unchanged.

Our SWE-Exp adaptation~\citep{chen2025sweexp} compiles the same five prior
attempts into an experience message. The coding agent receives this message
while generating the new trajectory, and the message remains in the exported
training record.

The SWE-Trail conditions keep prior attempts outside the coding agent's
context. H0 runs the same collection pipeline without history. H5 uses five
prior attempts, while H10 uses ten. Random detection replaces the
history-guided comparison location with a random eligible state while leaving
the remaining collection procedure unchanged. Random retention keeps H5's
history-guided comparison location but randomly chooses which executed route
continues. The randomized controls use the same per-trajectory collection
budget and execution horizon as H5.

\begin{table}[H]
	\centering
	\caption{Experimental conditions. Every condition produces 1,500 training
		records. The controlled adaptations use the same SeedBank attempts and
		training protocol as our main experiments.}
		\vspace{0.05in}
	\label{tab:condition-matrix-full}
	\small
	\setlength{\tabcolsep}{3pt}
	\renewcommand{\arraystretch}{1.1}
	\begin{tabularx}{\textwidth}{@{}
		>{\hsize=1.70\hsize\linewidth=\hsize\raggedright\arraybackslash}X
		>{\hsize=.78\hsize\linewidth=\hsize\raggedright\arraybackslash}X
		>{\hsize=.95\hsize\linewidth=\hsize\raggedright\arraybackslash}X
		>{\hsize=.90\hsize\linewidth=\hsize\raggedright\arraybackslash}X
		>{\hsize=.88\hsize\linewidth=\hsize\raggedright\arraybackslash}X
		>{\hsize=.79\hsize\linewidth=\hsize\raggedright\arraybackslash}X@{}}
			\toprule
			Condition & Source & \shortstack[l]{History visible\\to agent} & New collection & Training record & Role \\
			\midrule
			SeedBank
			& independent rollouts
			& no
			& complete rollout
			& generated trajectory
			& sampling baseline \\
			\addlinespace[2pt]
			
			SWE-Gym~\citep{pan2024swegym}, outcome-conditioned transform
			& SeedBank
			& no
			& no
			& outcome-conditioned record
			& outcome control \\
			\addlinespace[2pt]
			
			SWE-Lego~\citep{tao2026swelego}, strict tool-error masking
			& SeedBank
			& no
			& no
			& masked record
			& step control \\
			\addlinespace[2pt]
			
			SWE-TRACE~\citep{han2026swetrace}, critical-path reduction
			& SeedBank
			& no
			& no
			& compressed record
			& compression control \\
			\addlinespace[2pt]
			
			SWE-Exp~\citep{chen2025sweexp}, agent-visible experience
			& five attempts
			& yes
			& complete rollout
			& experience retained
			& visibility control \\
			\addlinespace[2pt]
			
			SWE-Trail-H0
			& no history
			& no
			& history-free collection
			& selected trajectory
			& no-history control \\
			\addlinespace[2pt]
			
			SWE-Trail-H5 (random detection)
			& five attempts
			& no
			& random comparison location
			& selected trajectory
			& location control \\
			\addlinespace[2pt]
			
			SWE-Trail-H5 (random retention)
			& five attempts
			& no
			& history-guided comparison
			& random executed route
			& retention control \\
			\addlinespace[2pt]
			
			\mainrow SWE-Trail-H5
			& five attempts
			& no
			& history-guided collection
			& selected trajectory
			& main method \\
			\addlinespace[2pt]
			
			SWE-Trail-H10
			& ten attempts
			& no
			& history-guided collection
			& selected trajectory
			& history-scale control \\
			\bottomrule
	\end{tabularx}
\end{table}

\section{Training, Evaluation, and Metrics}
\label{app:training}

\subsection{Controlled SFT Protocol}
\label{app:sft-protocol}

The primary student is initialized from Ling-3.0-flash-base, a
124B-parameter base model with 5.1B activated non-embedding parameters
\citep{inclusionai2026ling3flashbase}. To compare the training value of
different trajectory sets, we keep the remaining training setup fixed across
conditions. Table~\ref{tab:training-config} summarizes these controls.

Each main training corpus contains 1,500 coding trajectories. Every trajectory
is used whether or not it resolves the issue. Process-valid and
verified-positive subsets are reported only as diagnostics and do not determine
which trajectories enter training.

Each condition is trained independently three times from the same
initialization on 128 H800 GPUs using the same training schedule.
No condition receives separate hyperparameter tuning, and evaluation results
are never used to select a trained model. Each resulting model is evaluated
once on every task under the fixed protocol in
Appendix~\ref{app:evaluation}.

\textbf{Second student.}
We repeat the comparison with Qwen3-14B-Base~\citep{qwenteam2025qwen3base}.
The model is trained with full-parameter supervised fine-tuning on the
SeedBank, H0, H5, and H10 trajectory sets. Within this student, all training
settings other than the coding trajectories are again fixed. Each dataset is
trained independently three times, and every resulting model is evaluated once
under the same benchmark protocol. Appendix~\ref{app:second-student} reports
the complete results.

\begin{table}[!htbp]
	\centering
	\caption{Shared training settings for the 124B student. All listed settings
		are fixed across trajectory conditions.}
		\vspace{0.05in}
	\label{tab:training-config}
	\small
	\begin{tabular}{@{}ll@{}}
		\toprule
		Field & Controlled setting \\
		\midrule
		Initialization & Ling-3.0-flash-base \\
		Model scale & 124B total; 5.1B activated non-embedding parameters \\
		Coding trajectories & 1,500 per condition; all outcomes retained \\
		Background mixture & identical \\
		Tokenizer and tool representation & identical \\
		Context length and packing & identical \\
		Optimizer and learning-rate schedule & identical \\
		Update budget, passes, and precision & identical \\
		Randomization policy & identical \\
		Condition difference & coding trajectories only \\
		Agent reasoning supervision & identically enabled \\
		Model-selection rule & identical and fixed before evaluation \\
		Training repeats & three independent runs per condition \\
		Evaluation repeats & one complete evaluation per trained model \\
		Reported score & mean resolved count across the three runs \\
		\bottomrule
	\end{tabular}
\end{table}

\subsection{Evaluation Protocol}
\label{app:evaluation}

Within each benchmark, all students use the same evaluation setup. This includes
the autonomous OpenHands scaffold and prompt, available tools, sampling policy,
context and interaction budgets, repository construction, and containerized
verifier. The trained student is the only object that varies across conditions.
No condition receives inference-time tuning.

Every task remains in the denominator, including errors and timeouts under
their recorded outcomes. Per-run percentages are rounded only for display.
The reported condition mean is computed from the underlying resolved counts
before rounding.

For paired comparisons, each evaluation task contributes its mean binary
outcome across the three independently trained students in a condition.
We resample complete tasks with 20,000 bootstrap replicates and report the
percentile 95\% interval for the paired difference. We also count tasks for
which the first condition resolves more, fewer, or the same number of the
three trained models. These intervals describe uncertainty across evaluation
tasks conditional on the three trained models. Variation across training runs
is reported separately in Appendix~\ref{app:per-run}.
Table~\ref{tab:paired-student-history} in
Appendix~\ref{app:paired-contrasts} gives the complete paired contrasts.

\subsection{Metric Definitions}
\label{app:metrics}

This section defines the metrics used in Section~\ref{sec:results} and the
appendix tables.

\textbf{Outcome and collection yield.}
A trajectory is resolved when its final patch passes the benchmark tests.
Resolution is the fraction of evaluated tasks that are resolved.
A collection issue is counted as solved when at least one of its five newly
collected trajectories resolves. A verified positive is a resolved,
process-valid, nonempty selected trajectory that also passes the export checks.
The focused Top-5 analysis in Appendix~\ref{app:bon-audit} uses the same exact
strict-positive definition for both candidate banks.

\textbf{Process.}
Repeated reads count repeated accesses to the same normalized file target.
Validation counts executable checks. Post-final-edit validation records
whether at least one executable check occurs after the final edit.

\textbf{Collection mechanism.}
Candidate sampling counts a rollout in which a detector rule matches the
current state and the collector samples next-action candidates. Sampling alone
does not imply that a comparison is executed. Equivalent proposals or a failed
workspace restoration can prevent execution.

An executed comparison requires both selected candidate suffixes to run from
the same checkpoint. Candidate diversity is the fraction of candidate-sampling
cases in which the eight proposals contain at least two semantically distinct
actions. Alternative selection is the fraction of completed comparisons in
which the non-default route is retained.

Semantic progress records whether the selector finds clear evidence that one
executed suffix makes better local progress under the comparison objective
$u_i(d)$. We report this quantity over all rollouts. Candidate coverage asks
whether a useful action appears among the eight sampled proposals at all.
Appendix~\ref{app:selector-replay} evaluates this with exact-prefix replay.

\textbf{Tokens and time.}
Token counts use provider-reported prompt and completion tokens. Reported
reasoning tokens are a subset and are not added a second time.
Collection-overhead usage includes compiler, candidate-generation, additional
suffix, selector, and retry calls while excluding the ordinary coding rollout.
Where tables use the term \emph{acquisition-only}, it refers to this same
boundary. Summed trajectory-hours add elapsed time across individual tasks and
are not concurrent suite wall-clock time.

\textbf{Statistical tests for collection outcomes.}
The five trajectories collected for one issue are not independent observations,
so collection-level bootstrap intervals resample whole issues. Win/loss/tie
counts compare the number of resolved trajectories out of five for each issue.
Appendix~\ref{app:regime-records} additionally reports an exact sign-flip test
over nonzero issue-level differences. These collection-level analyses describe
differences in the collected data. The downstream training endpoint is reported
separately in Section~\ref{sec:student-results}.

\section{Complete Downstream Results}
\label{app:downstream-results}

\subsection{Per-run resolution}
\label{app:per-run}

Table~\ref{tab:per-run-results} reports all three independent training runs for
the 124B student. Every H5 run exceeds every SeedBank run on both benchmarks.
On Multilingual, the lowest H5 result, 64.00, also exceeds the highest H0
result, 63.67. The H5 and H0 ranges overlap on SWE-Pro.

For the H5--SeedBank comparison in Section~\ref{sec:student-results}, we use an
exact one-sided permutation test over the six independently trained models.
The test statistic is the difference in mean resolution,
\[
T=\bar{s}_{\mathrm{H5}}-\bar{s}_{\mathrm{SeedBank}}.
\]
Under the null hypothesis that the six run-level scores are exchangeable
between conditions, there are $\binom{6}{3}=20$ ways to assign three runs to
each condition. The exact $p$-value is the fraction of assignments whose mean
difference is at least $T_{\mathrm{obs}}$. On both benchmarks, the three H5
scores are the three highest scores, so only the observed assignment reaches
$T_{\mathrm{obs}}$. This gives a one-sided $p=1/20=0.05$ on each benchmark.
With three runs per condition, this is the smallest value attainable by this
test.

This test uses independently trained models as the statistical unit.
Appendix~\ref{app:paired-contrasts} examines differences across evaluation tasks
while keeping the trained models fixed.

\begin{table}[!htbp]
	\centering
	\caption{Independent downstream training runs, three per condition. Means
		are computed from integer resolved counts before rounding.
		Table~\ref{tab:condition-matrix-full} defines the experimental conditions.}
	\vspace{0.05in}
	\label{tab:per-run-results}
	\small
	\resizebox{\textwidth}{!}{%
		\begin{tabular}{@{}lrrrrrrrr@{}}
			\toprule
			& \multicolumn{4}{c}{Multilingual} & \multicolumn{4}{c}{SWE-Pro} \\
			\cmidrule(lr){2-5}\cmidrule(lr){6-9}
			Condition & R1 & R2 & R3 & Mean & R1 & R2 & R3 & Mean \\
			\midrule
			SeedBank independent rollouts & 63.00 & 63.67 & 63.33 & 63.33 & 41.18 & 41.31 & 41.18 & 41.22 \\
			SWE-Gym (outcome-conditioned transform) & 63.67 & 65.00 & 63.33 & 64.00 & 40.36 & 41.59 & 40.63 & 40.86 \\
			SWE-TRACE (critical-path reduction) & 65.33 & 64.67 & 63.33 & 64.44 & 41.72 & 40.08 & 40.49 & 40.77 \\
			SWE-Lego (strict tool-error masking) & 66.00 & 63.00 & 64.67 & 64.56 & 39.81 & 42.13 & 42.95 & 41.63 \\
			SWE-Exp (agent-visible experience) & 62.00 & 61.00 & 63.00 & 62.00 & 38.44 & 38.58 & 38.30 & 38.44 \\
			SWE-Trail-H0 & 63.67 & 63.00 & 63.67 & 63.44 & 41.04 & 41.72 & 43.09 & 41.95 \\
			SWE-Trail-H5 (random detection) & 64.67 & 63.33 & 64.67 & 64.22 & 41.59 & 41.72 & 41.31 & 41.54 \\
			SWE-Trail-H5 (random retention) & 66.00 & 64.67 & 63.67 & 64.78 & 41.04 & 42.13 & 41.72 & 41.63 \\
			\mainrow SWE-Trail-H5 & 67.67 & 65.33 & 64.00 & 65.67 & 41.72 & 43.37 & 43.37 & 42.82 \\
			SWE-Trail-H10 & 65.67 & 65.00 & 67.00 & 65.89 & 43.64 & 42.68 & 42.54 & 42.95 \\
			\bottomrule
	\end{tabular}}
\end{table}

\subsection{Task-paired contrasts}
\label{app:paired-contrasts}

We next examine downstream differences on the same evaluation tasks while
keeping the trained models fixed. The contrasts cover H5--H0, H10--H0,
H10--H5, and the stricter-selector condition of
Appendix~\ref{app:selector-replay} against H5.
Table~\ref{tab:paired-student-history} and
Figure~\ref{fig:paired-history-effects} use the task-paired bootstrap protocol
defined in Appendix~\ref{app:evaluation}. Comparisons with SeedBank use the
run-level results and mean differences in Appendix~\ref{app:per-run}.

The H5--H0 point estimates are positive on both benchmarks, at 2.22 points on
Multilingual and 0.87 points on SWE-Pro. Their 95\% task-bootstrap intervals,
however, both include zero. Increasing the history from five to ten attempts
adds only 0.22 and 0.14 points over H5. The stricter-selector condition is
below H5 on both benchmarks.

\begin{table}[!htbp]
	\centering
	\caption{Task-paired downstream contrasts conditional on the trained models.
		W/L/T counts tasks for which the first condition resolves more, fewer, or the
		same number of the three independently trained models. Intervals resample
		evaluation tasks rather than training runs.}
	\label{tab:paired-student-history}
	\small
	\resizebox{\textwidth}{!}{%
		\begin{tabular}{@{}lrlrl@{}}
			\toprule
			Contrast & Multilingual $\Delta$ [95\% CI] & W/L/T & SWE-Pro $\Delta$ [95\% CI] & W/L/T \\
			\midrule
			\mainrow H5 $-$ H0 & 2.22 [\textminus0.2, 4.6] & 55/35/210 & 0.87 [\textminus0.8, 2.5] & 130/111/490 \\
			H10 $-$ H0 & 2.44 [\textminus0.1, 4.9] & 61/39/200 & 1.00 [\textminus0.7, 2.7] & 134/112/485 \\
			H10 $-$ H5 & 0.22 [\textminus2.0, 2.5] & 51/49/200 & 0.14 [\textminus1.3, 1.6] & 117/114/500 \\
			Stricter selector $-$ H5 & \textminus1.78 [\textminus4.0, 0.4] & 42/58/200 & \textminus0.91 [\textminus2.6, 0.8] & 105/125/501 \\
			\bottomrule
	\end{tabular}}
\end{table}

\begin{figure}[!htbp]
	\centering
	\includegraphics[width=\textwidth]{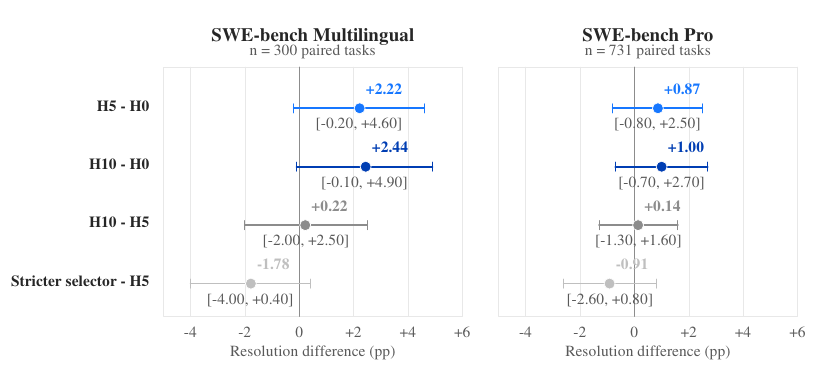}
	\caption{Task-paired downstream contrasts conditional on the trained models.
		Points show mean differences across the three trained models, and bars show
		task-cluster 95\% bootstrap intervals.}
	\label{fig:paired-history-effects}
\end{figure}

\subsection{Multilingual results by language}
\label{app:per-language}

Table~\ref{tab:per-language} breaks down Multilingual resolution by repository
language using the same three-run aggregation as the overall results. Each
language contains only 12--44 tasks, so the breakdown is descriptive.

H5 is no lower than H0 on seven of the nine languages. The two decreases, on
JavaScript and PHP, correspond to roughly one third of a task per run each.
The largest increase over H0 occurs on C.
%
\begin{table}[!htbp]
	\centering
	\caption{Multilingual resolution by repository language, in percent.
		$\Delta$ columns compare H5 with SeedBank and H0.}
	\vspace{0.05in}
	\label{tab:per-language}
	\small
	\setlength{\tabcolsep}{5pt}
	\begin{tabular}{@{}lrrrrrr@{}}
		\toprule
		Language & Tasks & SeedBank & H0 & H5 & $\Delta$ SeedBank & $\Delta$ H0 \\
		\midrule
		C & 30 & 60.0 & 56.7 & 68.9 & \tablegain{+8.9} & \tablegain{+12.2} \\
		C++ & 12 & 63.9 & 69.4 & 69.4 & \tablegain{+5.6} & 0.0 \\
		Go & 42 & 34.1 & 36.5 & 37.3 & \tablegain{+3.2} & \tablegain{+0.8} \\
		Java & 43 & 72.9 & 72.1 & 72.9 & 0.0 & \tablegain{+0.8} \\
		JavaScript & 31 & 74.2 & 72.0 & 71.0 & \tabledrop{\textminus3.2} & \tabledrop{\textminus1.1} \\
		PHP & 43 & 69.8 & 72.9 & 72.1 & \tablegain{+2.3} & \tabledrop{\textminus0.8} \\
		Ruby & 44 & 61.4 & 62.9 & 64.4 & \tablegain{+3.0} & \tablegain{+1.5} \\
		Rust & 43 & 74.4 & 71.3 & 75.2 & \tablegain{+0.8} & \tablegain{+3.9} \\
		TypeScript & 12 & 55.6 & 55.6 & 61.1 & \tablegain{+5.6} & \tablegain{+5.6} \\
		\midrule
		\mainrow All & 300 & 63.33 & 63.44 & 65.67 & \tablegain{+2.33} & \tablegain{+2.22} \\
		\bottomrule
	\end{tabular}
\end{table}

We also compare this pattern with the language composition of the collection
set. It contains only three C issues out of 300 and no Ruby issues,
while nearly one third of its issues are in Python, which does not appear in
Multilingual. The observed gains do not simply follow the language
composition of the collection set.

\subsection{Results for the Qwen3-14B Student}
\label{app:second-student}

We repeat the SeedBank, H0, H5, and H10 comparison with Qwen3-14B-Base to test
whether the downstream gains also appear with another student model.
Table~\ref{tab:second-student} reports three independent training runs under
the protocol of Appendix~\ref{app:sft-protocol}.

H5 improves mean resolution over SeedBank by 3.11 percentage points on
Multilingual and 2.42 points on SWE-Pro. It also exceeds H0 by 1.67 and
1.60 points, respectively. Every H5 run exceeds every SeedBank and H0 run on
both benchmarks.

Increasing the history from five to ten attempts adds 1.00 point on
Multilingual and 0.36 points on SWE-Pro. Every H10 run exceeds every H5 run on
both benchmarks. These additional gains are smaller than the H5--H0 gains.

\begin{table}[!htbp]
	\centering
	\caption{Resolution of the Qwen3-14B-Base student in three independent
		training runs per trajectory set. Values are percentages over the 300
		Multilingual and 731 SWE-Pro tasks. Means are computed from resolved counts
		before rounding, and $\Delta$ is relative to SeedBank in percentage points.}
	\vspace{0.05in}
	\label{tab:second-student}
	\small
	\resizebox{\textwidth}{!}{%
		\begin{tabular}{@{}lrrrrrrrrrr@{}}
			\toprule
			& \multicolumn{5}{c}{Multilingual} & \multicolumn{5}{c}{SWE-Pro} \\
			\cmidrule(lr){2-6}\cmidrule(lr){7-11}
			Training dataset & R1 & R2 & R3 & Mean & $\Delta$ & R1 & R2 & R3 & Mean & $\Delta$ \\
			\midrule
			SeedBank & 11.67 & 11.00 & 11.33 & 11.33 &  & 10.94 & 11.22 & 11.08 & 11.08 &  \\
			\textsc{SWE-Trail-H0} & 12.67 & 13.00 & 12.67 & 12.78 & +1.44 & 11.90 & 11.76 & 12.04 & 11.90 & +0.82 \\
			\mainrow \textsc{SWE-Trail-H5} & 14.33 & 14.67 & 14.33 & 14.44 & +3.11 & 13.27 & 13.54 & 13.68 & 13.50 & +2.42 \\
			\textsc{SWE-Trail-H10} & 15.33 & 15.33 & 15.67 & 15.44 & +4.11 & 13.82 & 13.95 & 13.82 & 13.86 & +2.78 \\
			\bottomrule
	\end{tabular}}
\end{table}

\section{Collection Mechanism in Detail}
\label{app:extended-results}
\label{app:regime-records}

\subsection{Actions, Repeated Reads, and Validation}
\label{app:process-distributions}

\begin{figure}[!htbp]
	\centering
	\includegraphics[width=\textwidth]{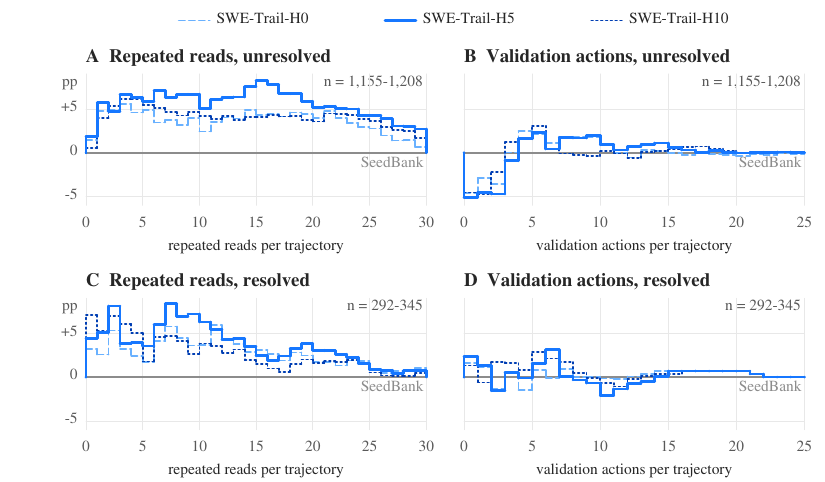}
	\caption{\textbf{H5 shows the largest shift toward fewer repeated reads,
			while validation counts remain close to SeedBank.}
		At each count, the curve gives the share of trajectories at or below that
		count minus the corresponding SeedBank share, in percentage points.
		SeedBank is the zero line. A positive value means that more trajectories
		fall at or below the displayed count. Panels A and C show repeated reads.
		H5 reaches 8.2 points above SeedBank for unresolved trajectories and
		8.4 points for resolved trajectories, compared with 5.6 and 5.9 for H0.
		Panels B and D show validation actions, where all curves remain within
		about five points of SeedBank. The rightmost bin includes values at or
		above the displayed cap. Coding-action distributions are omitted for
		legibility.}
	\label{fig:corpus-quality-ecdf}
\end{figure}

We first examine how the collected trajectories differ in their repair process,
not only in their final outcomes. To avoid attributing differences only to a
change in the resolved/unresolved mixture, we compare resolved and unresolved
trajectories separately. Figure~\ref{fig:corpus-quality-ecdf} reports
repeated-read and validation distributions relative to SeedBank.

Table~\ref{tab:corpus-process-means} complements these outcome-stratified
distributions with means over the full corpus. It also reports whether an
executable check follows the final edit, rather than the number of validation
actions.
Mean actions and repeated reads decrease from SeedBank to H0 and again from
H0 to H5, while post-final-edit validation increases. H10 partially reverses
each of these changes.

\begin{table}[!htbp]
	\centering
	\caption{Process means over all 1,500 collected trajectories per condition.
		Post-edit validation is the share of edited trajectories whose final edit
		is followed by an executable check.}
	\label{tab:corpus-process-means}
	\vspace{0.05in}
	\small
	\begin{tabular}{@{}lrrr@{}}
		\toprule
		Condition & Mean actions & Mean repeated reads & Post-edit validation \\
		\midrule
		SeedBank & 43.37 & 8.78 & 86.62\% \\
		\textsc{SWE-Trail-H0} & 40.89 & 7.59 & 88.91\% \\
		\mainrow \textsc{SWE-Trail-H5} & 39.51 & 7.01 & 89.53\% \\
		\textsc{SWE-Trail-H10} & 40.72 & 7.46 & 88.65\% \\
		\bottomrule
	\end{tabular}
\end{table}

\subsection{Student Behavior on Held-Out Issues}
\label{app:student-eval-behavior}

The preceding analysis describes the trajectories used for training. We next
ask whether similar process differences appear when the trained students solve
new issues. We compare evaluation trajectories from the 124B students trained
on H0 and H5 on both Multilingual and SWE-Pro. All evaluation tasks are
included, including failed and incomplete executions.

Coding actions are tool calls other than planning and finish calls. Repeated
reads count file-read occurrences beyond the first read of each normalized
target, including reads after an edit. Repeated-read share divides this count
by the number of recognized file-read occurrences in the trajectory.
Post-final-edit checking is measured only for trajectories that contain an
edit and records whether a recognized executable check occurs after the final
edit, regardless of whether that check passes.

\begin{table}[!htbp]
	\centering
	\caption{Descriptive evaluation behavior of the 124B students. Metrics are
		computed separately for each independent training run and then averaged
		across the three runs per condition. $\Delta$ is H5 minus H0. Differences
		in rates are in percentage points.}
		\vspace{0.05in}
	\label{tab:student-eval-behavior}
	\small
	\setlength{\tabcolsep}{4pt}
	\begin{tabular}{@{}lrrrrrr@{}}
		\toprule
		& \multicolumn{3}{c}{Multilingual} & \multicolumn{3}{c}{SWE-Pro} \\
		\cmidrule(lr){2-4}\cmidrule(l){5-7}
		Metric & H0 & H5 & $\Delta$ & H0 & H5 & $\Delta$ \\
		\midrule
		Actions / trajectory & 84.55 & 78.30 & \textminus6.25 & 77.84 & 75.34 & \textminus2.50 \\
		Repeated reads / trajectory & 16.41 & 14.89 & \textminus1.52 & 15.17 & 14.52 & \textminus0.65 \\
		Repeated-read share (\%) & 48.12 & 46.61 & \textminus1.51 & 44.77 & 44.17 & \textminus0.61 \\
		Post-final-edit check rate (\%) & 81.43 & 80.57 & \textminus0.86 & 86.39 & 87.43 & +1.04 \\
		\bottomrule
	\end{tabular}
\end{table}

\begin{samepage}
	Table~\ref{tab:student-eval-behavior} shows that H5-trained students take
	fewer actions and repeat fewer file reads on both benchmarks. Repeated-read
	share is also lower. The decreases in actions and repeated reads follow the
	same direction as the training-corpus differences in
	Appendix~\ref{app:process-distributions}. Post-final-edit checking decreases
	on Multilingual but increases on SWE-Pro.
	These results describe differences in student behavior and do not identify
	the cause of the resolution gains.
\end{samepage}

\subsection{Collection Yield by History Type}
\label{app:source-regime-yield}

For collection yield, we compare H0 and H5 across the three SeedBank history
types.
Table~\ref{tab:h0-h5-source-regime-yield} breaks down the 16 additional
resolved trajectories and six additional solved issues by the three SeedBank
history types.

\begin{table}[!htbp]
	\centering
	\caption{H0/H5 collection yield by SeedBank history type. An issue is
		counted as solved when at least one of its five newly collected trajectories
		resolves.}
			\vspace{0.05in}
	\label{tab:h0-h5-source-regime-yield}
	\small
	\begin{tabular}{@{}lrrrrrr@{}}
		\toprule
		& \multicolumn{3}{c}{Resolved trajectories} & \multicolumn{3}{c}{Solved issues} \\
		\cmidrule(lr){2-4}\cmidrule(lr){5-7}
		History type & H0 & H5 & $\Delta$ & H0 & H5 & $\Delta$ \\
		\midrule
		Stable-solved & 176 & 176 & 0 & 37 & 37 & 0 \\
		Mixed & 130 & 143 & +13 & 40 & 45 & +5 \\
		Never-solved & 23 & 26 & +3 & 15 & 16 & +1 \\
		\midrule
		All & 329 & 345 & +16 & 92 & 98 & +6 \\
		\bottomrule
	\end{tabular}
\end{table}

Mixed histories account for 13 of the 16 additional resolved trajectories and
five of the six additional solved issues. H5 solves every mixed issue solved
by H0 and five additional ones. Never-solved histories account for the
remaining three additional resolved trajectories and one net additional solved
issue. Within this group, H5 solves nine issues missed by H0, while H0 solves
eight missed by H5. The gain on mixed histories therefore preserves H0's
solved-issue coverage, whereas the smaller never-solved gain comes from
different sets of solved issues.

With issues as clusters, the H5$-$H0 difference in resolved trajectories is
$+1.07$ percentage points, with a 95\% CI of $[-0.33, 2.47]$ and an exact
sign-flip $p=0.16$. The difference in solved issues is $+6$, or $+2.0$
percentage points, with a 95\% CI of $[-1.0, 5.0]$~pp and $p=0.29$.
These collection-level comparisons are descriptive and do not identify which
trajectories account for the downstream training difference in
Section~\ref{sec:student-results}.

\subsection{Runtime Comparison Stages}
\label{app:branch-funnel}

We next examine where the H0 and H5 collection processes differ.
Table~\ref{tab:branch-funnel-exact} expands
Figure~\ref{fig:branch-funnel} by history type and gives the denominator for
each stage.

Candidate sampling occurs in 664 H0 rollouts and 1,147 H5 rollouts.
At least two semantically distinct actions remain in 598 and 1,030 of these
cases, respectively. Both candidate suffixes are successfully executed in
560 H0 and 968 H5 rollouts. The non-default route is retained in 218 and
362 completed comparisons.

The largest difference occurs in how often rollouts reach candidate sampling.
Once candidates are sampled, H0 and H5 have similar candidate diversity,
execution rates, and alternative-selection rates. Because the two collectors
sample candidates at different points in their rollouts, similar conditional
rates do not imply that they generate the same candidate distributions.

\begin{table}[!htbp]
	\centering
	\caption{H0/H5 collection stages by SeedBank history type. H0 history labels
		are joined post hoc by issue ID and were never available during H0
		collection. Candidate diversity is conditional on candidate sampling,
		executed comparison is conditional on candidate diversity, and alternative
		selection is conditional on a completed comparison. Candidate sampling,
		semantic progress, and resolution use all trajectories.}
	\label{tab:branch-funnel-exact}
	\small
	\resizebox{\textwidth}{!}{%
		\begin{tabular}{@{}llrrrrrrr@{}}
			\toprule
			Condition & History type & Traj. &
			\shortstack{Candidate\\sampling} &
			\shortstack{Distinct\\actions} &
			\shortstack{Executed\\comparison} &
			Alt. selected & Semantic progress & Resolved \\
			\midrule
			H0 & stable-solved & 185 & 36.22\% & 91.04\% & 86.89\% & 30.19\% & 21.08\% & 95.14\% \\
			\mainrow H5 & stable-solved & 185 & 53.51\% & 90.91\% & 92.22\% & 43.37\% & 35.14\% & 95.14\% \\
			H0 & mixed & 255 & 53.73\% & 93.43\% & 95.31\% & 43.44\% & 35.29\% & 50.98\% \\
			\mainrow H5 & mixed & 255 & 83.53\% & 94.37\% & 94.53\% & 41.05\% & 56.47\% & 56.08\% \\
			H0 & never-solved & 1,060 & 43.40\% & 88.91\% & 94.13\% & 38.80\% & 28.49\% & 2.17\% \\
			\mainrow H5 & never-solved & 1,060 & 78.77\% & 88.50\% & 94.05\% & 35.68\% & 37.92\% & 2.45\% \\
			H0 & all & 1,500 & 44.27\% & 90.06\% & 93.65\% & 39.00\% & 28.73\% & 21.93\% \\
			\mainrow H5 & all & 1,500 & 76.47\% & 89.80\% & 93.98\% & 37.40\% & 40.73\% & 23.00\% \\
			\bottomrule
	\end{tabular}}
\end{table}

\subsection{Compiler Classes and Representative Records}
\label{app:compiler-records}

The SeedBank history types above describe only the outcomes of the five prior
attempts. The compiler separately interprets the evidence in each history and
may abstain when that evidence does not support additional collection effort.
Table~\ref{tab:h5-skill-regime} reports these compiler classes and the number
of detector rules admitted for collection.

Among histories assigned to the stable-solved, mixed, and never-solved
compiler classes, mixed and never-solved histories have more active rules per
issue, at 1.61 and 1.47, than stable-solved histories at 0.62. The compiler
also abstains more often on stable-solved histories. Twelve issues are assigned
to the insufficient-evidence class, for which no rule is admitted.

\begin{table}[!htbp]
	\centering
	\caption{H5 acquisition-record statistics by compiler class. Per-class means
		are rounded to two decimals. Appendix~\ref{app:detector-schema} gives the
		exact totals of 687 compiled and 397 active detector rules.}
		\vspace{0.05in}
	\label{tab:h5-skill-regime}
	\small
	\begin{tabular}{@{}lrrrr@{}}
		\toprule
		Compiler class & Issues & Mean rules & Active rules & Abstention \\
		\midrule
		Stable-solved & 37 & 2.38 & 0.62 & 37.84\% \\
		Mixed & 41 & 2.22 & 1.61 & 7.32\% \\
		Never-solved & 210 & 2.29 & 1.47 & 13.81\% \\
		Insufficient & 12 & 2.33 & 0.00 & 100.00\% \\
		\bottomrule
	\end{tabular}
\end{table}

\textbf{Representative compiler records.}
The following records illustrate three different compiler outcomes: retaining
uncertainty around an otherwise successful history, allocating additional
collection effort when competing repair choices remain, and abstaining when
the available evidence is insufficient. The never-solved example in
Appendix~\ref{app:rescue-case} follows one admitted rule through online
execution. Predicates and objectives below are reproduced verbatim.
Omitted fields are marked with \texttt{\ldots}. The status
\texttt{active} means that the rule passed compiler admission. It does not
indicate that the rule matched a new rollout. All objectives remain private
to the collector.

\begin{sweartifact}{Stable-solved \textbar\ shepmaster\_\_snafu-237}
	source outcomes: 5/5 resolved\par
	open uncertainty u1: whether other files in the snafu-derive crate contain
	quote! blocks with unhygienic std:: or core:: references, since the probes
	inspected only src/lib.rs\par
	frozen predicate: has\_repository\_evidence = true AND unique\_target\_count >= 4
	AND has\_validation = false AND last\_category = "search"\par
	objective: Preserve both viable editing routes until commitment; avoid
	premature route elimination\par
	status: active (1 of 2 proposed states activated; the other stays dormant)
\end{sweartifact}

\begin{sweartifact}{Mixed \textbar\ cloudflare\_\_cloudflare-go-1066}
	source outcomes: 4/5 resolved\par
	competing hypotheses: h1 adopt the repository's existing custom UnmarshalJSON
	pattern; h2 change PctComplete from int to string with no custom
	unmarshaling\par
	frozen predicate: has\_repository\_evidence = true AND has\_edit = false AND
	validation\_count = 0 AND search\_count >= 3 AND semantic\_phase =
	"hypothesis"\par
	objective: Prefer proceeding to minimal type-change edit over continued search
	for custom unmarshaling patterns\par
	status: active (3 of 3 proposed states activated)
\end{sweartifact}

\begin{sweartifact}{Insufficient \textbar\ adobe\_\_elixir-styler-203}
	source outcomes: 0/5 resolved\par
	open uncertainties: u1 whether the Zipper module exposes parent-node access;
	u2 whether any existing style module demonstrates parent-context detection;
	u3 the AST shape of first..last//step \ldots\par
	proposed predicate: edit\_count = 0 AND recent10\_read\_search\_count >= 4 AND
	action\_count >= 6\par
	objective: Prefer implementing the template-driven easy task (List.zip) over
	continued exploration or entangled multi-task planning\par
	status: dormant (0 of 2 proposed states activated; the compiler abstains)
\end{sweartifact}

\section{Component Controls, Cost, and \texorpdfstring{Best-of-$N$}{Best-of-N}}
\label{app:ablations}

\subsection{Randomized Controls and Selector Replay}
\label{app:selector-replay}

The two randomized controls test where comparisons occur and which executed
route continues. Both use H5's five-attempt history, sampling-temperature
schedule, training protocol, and evaluation protocol. They also use the same
per-trajectory collection budget and execution horizon as H5. This budget
allows eight candidate proposals, one shortlist call, one selector call, and
at most two two-observation suffix executions. Realized token counts may still
differ across trajectories.

\textbf{Random detection.}
Random detection replaces H5's history-guided comparison location with a
random eligible state while leaving the remaining collection procedure
unchanged. Eligible states are repository-observation boundaries at which the
observation has completed, the workspace can be restored, no earlier comparison
has been executed, and enough budget remains for both short suffixes. The state
is sampled uniformly from these eligible locations without using history or a
detector score. If a trajectory contains no eligible state, it completes as an
ordinary rollout without a comparison.

\textbf{Random retention.}
Random retention keeps H5's history-guided comparison location and executes the
same candidate-generation, shortlisting, and suffix-comparison procedure. Once
both suffixes execute successfully, a fair binary draw determines which
workspace continues, regardless of the selector's judgment. If restoration or
suffix execution fails, the default route continues.

\textbf{Separate collection with a stricter selector.}
We also test whether requiring stronger execution evidence improves the
training value of the collected trajectories. This condition runs a separate
H5-style collection in which an alternative route is preferred only when its
executed suffix provides stronger evidence about the root cause or working
hypothesis. The collection resolves 343 of 1,500 trajectories across 97 issues,
compared with 345 trajectories across 98 issues for H5. Students trained on
these trajectories reach 63.89\% on Multilingual and 41.91\% on SWE-Pro,
1.78 and 0.91 points below H5. Thus, the stricter criterion leaves collection
yield nearly unchanged but produces lower student means on both benchmarks.

\textbf{Fixed-comparison diagnostic replay.}
To examine the selector without changing candidate generation or execution, we
replay 18 H5 comparisons under the stricter criterion. The visible prefix,
workspace, candidates, and both executed suffixes remain fixed, so only the
selection judgment changes. These comparisons come from seven issues that the
compiler classifies as mixed.

The stricter criterion changes eight decision labels and six retained routes.
Some changed decisions become ties and therefore retain the default route,
which explains the difference between these counts. Ties increase from 5 of
18 comparisons to 12 of 18. The stricter criterion mainly converts uncertain
preferences into abstentions.

Candidate coverage is the most frequent non-exclusive limitation in this
sample, appearing in 13 of 18 comparisons.
Seven have partial coverage and six have poor coverage. Selection is identified
as a limitation in 4 of 18 comparisons, followed by suffix horizon and
post-selection drift at 3 of 18 each. A selector cannot recover a useful action
that was not proposed, which motivates testing more diverse candidate proposals.
These counts describe the 18 replayed comparisons rather than all 968 executed
comparisons.

\subsection{History Scale and Collection Cost}
\label{app:cost}

We compare H0, H5, and H10 to measure how the benefit and cost change as more
history is provided. Costs cover new collection given existing prior attempts,
under the accounting boundary in Appendix~\ref{app:cost-boundary}.
Figure~\ref{fig:history-scale-dashboard} combines the
recorded collection cost with downstream results for the 124B student.

\begin{figure}[!htbp]
	\centering
	\includegraphics[width=\textwidth]{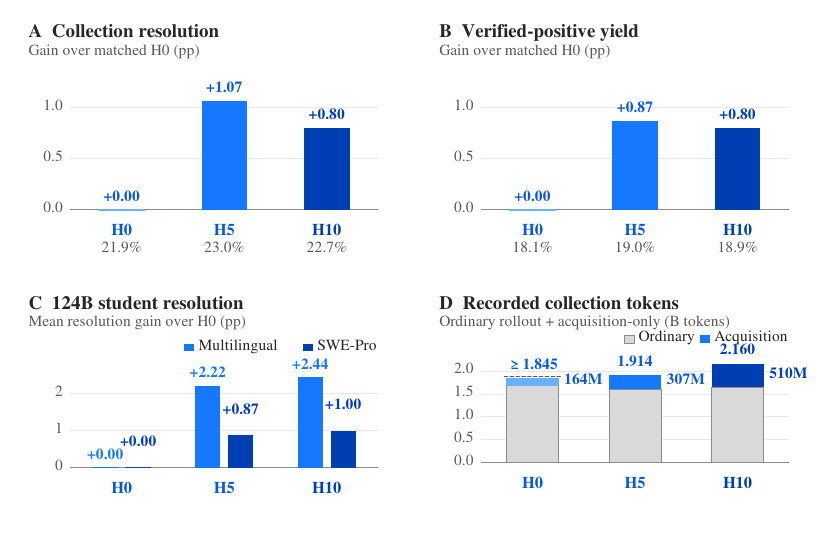}
	\caption{\textbf{H10 adds little mean improvement over H5 while collection
			cost continues to increase.} Collection outcomes and 124B-student gains are
		shown relative to H0. Recorded tokens are divided into ordinary rollout and
		acquisition-only usage. H0 is a lower bound because its offline compiler
		calls were not recorded.}
	\label{fig:history-scale-dashboard}
\end{figure}

H5 records 1.914B total collection tokens, including 307.3M acquisition-only
tokens. H0 records at least 1.845B total tokens, including 164.3M
acquisition-only runtime tokens. Its offline compiler tokens were not recorded.
Relative to this lower bound, H5 uses at most 3.7\% more total tokens than H0.
Removing acquisition-only usage leaves 1.607B ordinary rollout tokens for H5
and 1.681B for H0, a 4.4\% reduction.

H10 uses 2.160B total tokens, including 510.0M acquisition-only tokens.
These are 12.9\% and 66.0\% above H5, respectively. Ordinary rollouts use
1.650B tokens.

To show where H5's 307.3M acquisition-only tokens are spent, we break them
down by component and history type. Candidate proposals use 238.4M tokens,
executed suffixes 43.1M, selection 16.8M, and offline compilation 9.1M.
Across history types, stable-solved histories use 22.32M tokens, or 7.3\%.
Mixed histories use 51.97M, or 16.9\%, and never-solved histories use 232.85M,
or 75.8\%.

Table~\ref{tab:environment-totals} gives the corresponding environment-side
costs. Trajectory-hours sum elapsed task time and do not represent concurrent
wall-clock time.

\begin{table}[!htbp]
	\centering
	\caption{Environment-side totals over the 1,500 collected trajectories in
		each condition.}
		\vspace{0.05in}
	\label{tab:environment-totals}
	\small
	\begin{tabular}{@{}lrrr@{}}
		\toprule
		Condition & Coding actions & Validations & Trajectory-hours \\
		\midrule
		SeedBank & 65,056 & 6,814 & 355.2 \\
		\textsc{SWE-Trail-H0} & 61,329 & 6,760 & 196.2 \\
		\mainrow \textsc{SWE-Trail-H5} & 59,260 & 6,700 & 239.4 \\
		\textsc{SWE-Trail-H10} & 61,085 & 6,721 & 208.1 \\
		\bottomrule
	\end{tabular}
\end{table}

\subsection{Verifier-Based Training Corpus Selection}
\label{app:bon-audit}

We next test whether post-hoc selection can replace history-guided collection
or complement it. The first comparison selects from ten independently sampled
SeedBank trajectories per issue. The second applies the same selection rule to
five SeedBank and five H5 trajectories. Both retain five trajectories per
issue for student training.

\textbf{Ranking rule.}
For each issue, the verifier ranks ten candidates and retains five.
Format-admissible candidates rank ahead of inadmissible candidates, followed by
the binary SWE-rebench resolved label. Remaining ties use the same deterministic
SHA256-based rule in both comparisons. Reference patches, trajectory length,
and process metrics are not used for ranking.

\textbf{Budget and selected corpus.}
Top-5-of-10 uses twice as many independent rollouts as the five-rollout
SeedBank baseline. Its collection cost is not matched exactly to H5.
Both selection conditions rank 3,000 candidate trajectories and retain 1,500
for training.

\begin{table}[!htbp]
	\centering
	\caption{Verifier-based Top-5 selection. Both candidate sets contain 3,000
		trajectories and retain 1,500 trajectories for training.}
			\vspace{0.05in}
	\label{tab:hybrid-best-of-n}
	\small
	\setlength{\tabcolsep}{4pt}
	\begin{tabular}{@{}lrrrrr@{}}
		\toprule
		Candidate set & Res. & Solved issues & Strict $+$ & Never traj. & Never issues \\
		\midrule
		10 SeedBank & 394 & 99 & 313 & 15 & 11 \\
		\mainrow 5 SeedBank + 5 H5 & 413 & 104 & 321 & 26 & 16 \\
		\bottomrule
	\end{tabular}
\end{table}

Table~\ref{tab:hybrid-best-of-n} compares the two selected corpora.
The hybrid set has a 1.27 percentage-point higher trajectory-resolution rate
under issue-cluster pairing, with a 95\% CI of $[-0.07, 2.67]$ and a
29/17/254 win/loss/tie record. All disagreements in solved-issue coverage occur
on never-solved histories, with ten issues solved only by the hybrid set and
five only by the independent set. Both selected sets preserve coverage on
stable-solved and mixed histories. The selected banks show complementary issue
coverage under the same ranking rule.

\textbf{Student results.}
Table~\ref{tab:selection-students} reports students trained on the selected
trajectories under the protocol in Appendix~\ref{app:training}.

\begin{table}[!htbp]
	\centering
	\caption{Student resolution after verifier-based selection. Each corpus has
		three independent training runs. Ranges give the lowest and highest run.
		SeedBank and H5 repeat the results in Table~\ref{tab:main-sft-results}.}
			\vspace{0.05in}
	\label{tab:selection-students}
	\small
	\setlength{\tabcolsep}{5pt}
	\begin{tabular}{@{}lcccc@{}}
		\toprule
		& \multicolumn{2}{c}{Multilingual} & \multicolumn{2}{c}{SWE-Pro} \\
		\cmidrule(lr){2-3}\cmidrule(l){4-5}
		Trajectory corpus & Mean & Range & Mean & Range \\
		\midrule
		SeedBank & 63.33 & [63.00, 63.67] & 41.22 & [41.18, 41.31] \\
		Top-5-of-10 & 64.11 & [63.33, 65.00] & 41.86 & [41.45, 42.13] \\
		\textsc{SWE-Trail-H5} & 65.67 & [64.00, 67.67] & 42.82 & [41.72, 43.37] \\
		\mainrow Top-5-hybrid & 66.22 & [65.33, 66.67] & 43.14 & [42.82, 43.37] \\
		\bottomrule
	\end{tabular}
\end{table}

Top-5-of-10 improves over SeedBank but remains below H5 by 1.56 points on
Multilingual and 0.96 points on SWE-Pro. Thus, selecting from a larger pool of
independent trajectories does not recover the gains from SWE-Trail in this
comparison.

Top-5-hybrid reaches 66.22 on Multilingual and 43.14 on SWE-Pro, exceeding H5
by 0.56 and 0.32 points, respectively. These results show that post-hoc
selection can complement SWE-Trail when applied to a candidate set that
includes its trajectories.

\section{Case Study and Advice Diagnostics}
\label{app:cases}

\subsection{A Repair Case after Five Failed Attempts}
\label{app:rescue-case}

\begin{figure}[!htbp]
	\centering
	\includegraphics[width=\textwidth]{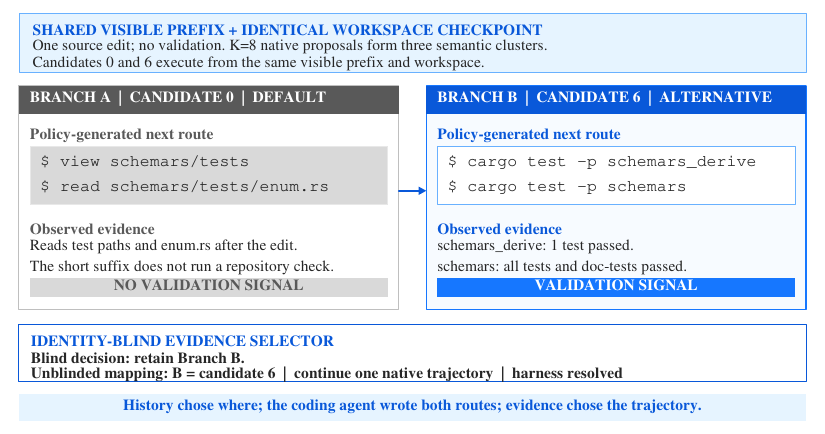}
	\caption{\textbf{History guides where additional collection effort is spent.}
		In the \texttt{schemars-304} example, both candidate suffixes are generated
		by the same coding agent from the same visible checkpoint. Candidate~0
		continues reading, while candidate~6 runs the relevant Rust tests. The
		selector uses the resulting repository evidence to choose which route
		continues.}
	\label{fig:schemars-walkthrough}
\end{figure}

This case shows how information from earlier failed attempts guides collection
without entering the coding agent's context. For
\texttt{gresau\_\_schemars-304}, all five SeedBank attempts fail after making
a one-line \texttt{SERDE\_KEYWORDS} edit without validating the change.
Under H5, four of five newly collected trajectories resolve the issue.
We follow one H5 trajectory from the compiled record through candidate
execution and route selection. Figure~\ref{fig:schemars-walkthrough} shows the
two executed suffixes.

\begin{swedossier}{\texttt{schemars-304} \textbar\ never-solved \textbar\ H5 attempt 4}
	
	\textbf{1. Earlier attempts identify where more collection effort may be useful.}
	All five prior attempts are unresolved. They repeatedly identify or edit
	\texttt{rename\_all\_fields}, then return to repository reading without
	validating the edit. A read-only repository check locates the
	\texttt{SERDE\_KEYWORDS} list that controls which serde attributes are
	forwarded, together with its definition and two uses. The compiled record
	keeps the sufficiency of the one-line edit as an open question.
	
	\begin{sweartifact}{Acquisition record}
		source outcome: 0/5 resolved\par
		uncertainty: is the edited keyword sufficient?\par
		probe: locate the keyword list that gates attribute forwarding\par
		observed: \texttt{SERDE\_KEYWORDS} definition and two uses, contents unobserved\par
		failure boundary: post-edit reading without validation
	\end{sweartifact}
	
	\textbf{2. The compiler turns this pattern into an observable rule.}
	The rule \texttt{post\_edit\_read\_loop\_without\_validation} matches after an
	edit when no subsequent validation has occurred and at least three of the ten
	most recent actions are reads or searches. When this pattern appears in the new
	rollout, the coding agent proposes eight next actions from its ordinary context.
	These proposals form three semantic groups.
	
	\tcblower
	
	\textbf{3. Short execution provides the evidence for route selection.}
	Candidate~0 continues reading the tests directory and \texttt{enum.rs}.
	Candidate~6 instead runs the two relevant Rust test suites from the same
	checkpoint.
	
	\begin{sweartifact}{Suffix evidence}
		A / candidate 0: no validation signal\par
		B / candidate 6: schemars\_derive -- 1 test passed\par
		B / candidate 6: schemars -- all tests and doc-tests passed\par
		blind selector: retain B\par
		unblinded mapping: B = candidate 6
	\end{sweartifact}
	
	The collector continues only from candidate~6's workspace. The exported record
	remains one continuous trajectory generated by the coding agent, and its
	one-file patch resolves under the SWE-rebench harness. History determines where
	additional collection effort is spent, while the coding agent itself proposes
	the test command that provides the deciding evidence.
	
\end{swedossier}

\subsection{Retaining and Removing Advice in Training Trajectories}
\label{app:advice-exports}

We separately examine whether removing advice from an advice-conditioned
trajectory changes its value for training. This experiment differs from the
SWE-Exp adaptation in Section~\ref{sec:student-results}. Here, reviewer-generated
advice is inserted into the coding agent's conversation during the rollout.

\textbf{Two exports of the same trajectories.}
We create two training corpora from the same 1,500 advice-conditioned
trajectories and train students under the protocol in
Appendix~\ref{app:training}. The two exports preserve trajectory membership and
order, coding actions, tool arguments, and tool results.

The advice-retained export keeps each aligned advice message and places it
before the agent-generated reasoning in the corresponding assistant tool-call
turn. The advice-removed export removes the advice messages and only the
reasoning fragments that explicitly attribute an action to that advice.
All remaining agent reasoning and actions are unchanged.
Both exports retain actions generated while advice was visible to the agent.
Removing advice changes the training record, not the context in which those
actions were generated.

\begin{table}[!htbp]
	\centering
	\caption{Three-run SFT means for two exports of the same advice-conditioned
		trajectories.}
	\label{tab:advice-export-sft}
	\small
	\begin{tabular}{@{}lrr@{}}
		\toprule
		Training corpus & Multilingual & SWE-Pro \\
		\midrule
		Independent SeedBank & 63.33 & 41.22 \\
		Advice removed & 63.00 & 42.23 \\
		Advice retained & 64.56 & 41.63 \\
		\mainrow SWE-Trail-H5 & 65.67 & 42.82 \\
		\bottomrule
	\end{tabular}
\end{table}
Table~\ref{tab:advice-export-sft} reports the resulting student means.
Retaining advice improves over SeedBank on both benchmarks. Removing it lowers
Multilingual resolution but raises SWE-Pro resolution relative to retaining it.
Advice removal does not consistently improve downstream performance in this
comparison.

\input{appendix/related_work}

%% file: appendix/related_work.tex
\section{Detailed Related Work}
\label{app:related}
Section~\ref{sec:related} groups the closest work into search and selection,
trajectory construction and curation, and experience-guided agents.
Here we give a broader comparison centered on two questions: who reads prior
attempts, and how those attempts affect later computation.
Table~\ref{tab:history-interfaces} summarizes the main interfaces.
\begin{table}[ht]
	\centering
	\caption{How prior attempts are reused. The main distinction is whether
		history enters the coding agent's context or remains private to an external
		collection procedure.}
		\vspace{0.05in}
	\label{tab:history-interfaces}
	\small
	\begin{tabular}{@{}p{0.21\textwidth}p{0.19\textwidth}p{0.30\textwidth}p{0.18\textwidth}@{}}
		\toprule
		Use of prior attempts & Consumer & Effect on later computation & Primary goal \\
		\midrule
		Retrieved reflection or memory
		& Coding agent
		& Adds previous feedback or experience to the visible context
		& Improve the current solve \\
		Distilled skill or workflow
		& Coding agent or workflow
		& Supplies reusable procedural guidance
		& Improve the current solve \\
		Inference-time replay or refinement
		& Search harness and coding agent
		& Reuses prefixes or conditions generation on prior summaries
		& Improve or select a current solution \\
		History-guided collection (ours)
		& Collector
		& Guides where additional collection effort is spent while history remains outside the agent context
		& Collect training trajectories \\
		\bottomrule
	\end{tabular}
\end{table}
\subsection{Executable Coding Agents and Trajectory Corpora}

SWE-bench established executable repository-level repair, while SWE-agent and
OpenHands provided widely used coding-agent scaffolds
\citep{jimenez2023swebench,yang2024sweagent,wang2024openhands}.
SWE-Gym, R2E-Gym, SWE-smith, and Open-SWE-Traces expand the available tasks,
environments, rollouts, and languages for training
\citep{pan2024swegym,jain2025r2egym,yang2025swesmith,ahmad2026openswetraces}.
These efforts increase the supply of coding trajectories.
Our focus is different: given several prior attempts on the same issue, we ask
how that history can guide the collection of additional trajectories.

\subsection{Inference-Time Search and Selection}

A common inference-time strategy samples several complete rollouts and selects
or combines them using tests, execution outcomes, or model judgments
\citep{ehrlich2025codemonkeys}.
Other methods reuse earlier computation more directly. They may restore a
previous trajectory prefix and continue from an earlier point, or summarize
prior attempts and use those summaries to condition later rollouts
\citep{ding2026swereplay,kim2026scalingagentic}.
These methods use additional computation to improve the solution returned for
the current issue.

Tree-search methods provide another form of adaptive inference.
MCTS and related approaches compare alternatives within a growing online search
tree and allocate computation using value estimates, process signals, or
terminal outcomes
\citep{kocsis2006bandit,zhou2023lats,antoniades2024swesearch}.
They can revisit a shared prefix and explore alternatives that independent
full rollouts may miss.

SWE-Trail differs in both source and objective. Its guidance comes from
independent attempts collected before the new rollout, and the goal is to
collect training trajectories rather than to return the best solution from a
search tree. It compiles fixed rules before collection and executes at most
one comparison during a rollout, without recursive tree expansion or value
backup. The H0 control in Section~\ref{sec:experiments} evaluates this bounded
collection procedure without history, rather than a full tree-search policy.

\subsection{Trajectory Construction and Post-Hoc Curation}

Several methods improve coding-agent training data after trajectories have
already been generated. SWE-Gym and related pipelines retain trajectories
based on executable outcomes, while other methods mask flawed tool calls,
remove redundant interactions, reconstruct better paths, or use process-level
signals to identify more useful supervision
\citep{pan2024swegym,tao2026swelego,ma2026patches,
	han2026swetrace,xiao2025agentdiet}.
Data valuation and training-dynamics methods similarly estimate the usefulness
of examples that already exist
\citep{koh2017influence,ghorbani2019datashapley,paul2021datadiet,
	xie2023doremi}.

These methods are complementary to SWE-Trail. Post-hoc curation decides which
parts of completed data should be retained or emphasized. SWE-Trail instead
uses prior attempts while a new trajectory is being collected. The comparison
can produce new repository evidence that would not exist if the alternative
action were never executed. Curation can still be applied afterward.
Appendix~\ref{app:bon-audit} evaluates one such combination using
verifier-based selection over SeedBank and H5 trajectories.

\subsection{Training-Time Diversity and Adaptive Curation}

Recent work treats the training corpus itself as an object to optimize.
TDScaling increases trajectory diversity through blueprint-driven synthesis
and adaptive task evolution rather than only sampling more trajectories
\citep{chen2026tdscaling}.
Han studies filtering of an existing SWE trajectory pool and reports an
interaction between data quality, quantity, and model scale
\citep{han2026trajectorycuration}.
CurateEvo represents curation as executable code and updates the curation
procedure from downstream failures to construct SFT, RL, and memory data from
a fixed raw corpus
\citep{wang2026curateevo}.

SWE-Trail shares the goal of improving the training distribution but acts at a
different stage. It changes how new trajectories are collected. Earlier
attempts on the same issue guide where additional collection effort is spent,
while the coding agent generates the candidate actions without receiving that
history.

\subsection{Experience-Guided Agents}

Another line of work converts previous interaction into information that helps
an agent act later. Reflexion and ExpeL store feedback or reusable experience
across attempts
\citep{shinn2023reflexion,zhao2023expel}.
Agent Workflow Memory, CODESKILL, Socratic-SWE, and ProPlay derive workflows,
skills, or procedural knowledge from previous executions
\citep{wang2024awm,li2026codeskill,xiao2026socraticswe,ma2026proplay}.
For software repositories, experience may come from earlier repair
trajectories, human debugging records, or trajectories produced during the
current solve
\citep{chen2025sweexp,lin2024evocoder,mu2025experepair,
	wang2026memgovern,hayashi2025sage,xia2025livesweagent}.
Other systems recombine agents or revise failed paths during inference
\citep{ehrlich2025codemonkeys,lin2025seagent}.

The common pattern is that distilled experience changes the coding agent's
visible context, workflow, or behavior. SWE-Trail keeps that information
outside the coding agent. History guides the collector in deciding where
additional collection effort may be useful, while the coding agent proposes
its next actions from the ordinary context. This separation is central to our
training-data setting because the exported trajectory does not depend on
history being available to the student at test time.

\subsection{Adaptive Acquisition and Harness Optimization}

The broader idea of allocating computation using previous observations appears
in active learning and experimental design, which choose informative examples
or experiments
\citep{sener2018coreset,kirsch2019batchbald,foster2021dad}.
Intrinsic-motivation methods similarly direct interaction toward novel or
prediction-reducing states
\citep{bellemare2016pseudocount,pathak2017curiosity}.
For LLM agents, direction-informed gating learns from counterfactual
exploration whether a state feature indicates that extra rollouts will help
or hurt~\citep{li2026same}, and paired rollouts can estimate the value of
asking a clarification question before acting~\citep{huang2026cleat}.
Agent self-improvement and harness-optimization methods adapt prompts,
workflows, tools, or control procedures from execution feedback
\citep{madaan2023selfrefine,xi2024agentgym,robeyns2025sica,
	pan2026rho,lin2026ahe}.

SWE-Trail applies a narrower form of adaptive allocation within one issue.
The history of that issue is compiled before the new collection round, and the
resulting rules remain fixed during the rollout. They guide where additional
collection effort is spent and how executed alternatives are compared, without
changing the coding-agent prompt, tools, or model weights.

Recent results show that simple search can remain a strong baseline
\citep{gideoni2026simple}. We compare against independent sampling,
a history-free version of the same collection pipeline, randomized controls,
and verifier-based selection in Section~\ref{sec:experiments}.

%% file: references.bib
@inproceedings{jimenez2023swebench,
  author       = {Carlos E. Jimenez and
                  John Yang and
                  Alexander Wettig and
                  Shunyu Yao and
                  Kexin Pei and
                  Ofir Press and
                  Karthik R. Narasimhan},
  title        = {SWE-bench: Can Language Models Resolve Real-world Github Issues?},
  booktitle    = {ICLR},
  year         = {2024}
}

@inproceedings{yang2024sweagent,
  author       = {John Yang and
                  Carlos E. Jimenez and
                  Alexander Wettig and
                  Kilian Lieret and
                  Shunyu Yao and
                  Karthik Narasimhan and
                  Ofir Press},
  title        = {SWE-agent: Agent-Computer Interfaces Enable Automated Software Engineering},
  booktitle    = {NeurIPS},
  year         = {2024}
}

@inproceedings{wang2024openhands,
  author       = {Xingyao Wang and
                  Boxuan Li and
                  Yufan Song and
                  Frank F. Xu and
                  Xiangru Tang and
                  Mingchen Zhuge and
                  Jiayi Pan and
                  Yueqi Song and
                  Bowen Li and
                  Jaskirat Singh and
                  Hoang H. Tran and
                  Fuqiang Li and
                  Ren Ma and
                  Mingzhang Zheng and
                  Bill Qian and
                  Yanjun Shao and
                  Niklas Muennighoff and
                  Yizhe Zhang and
                  Binyuan Hui and
                  Junyang Lin and
                  et al.},
  title        = {OpenHands: An Open Platform for {AI} Software Developers as Generalist
                  Agents},
  booktitle    = {ICLR},
  year         = {2025}
}

@inproceedings{pan2024swegym,
  author       = {Jiayi Pan and
                  Xingyao Wang and
                  Graham Neubig and
                  Navdeep Jaitly and
                  Heng Ji and
                  Alane Suhr and
                  Yizhe Zhang},
  title        = {Training Software Engineering Agents and Verifiers with SWE-Gym},
  booktitle    = {ICML},
  year         = {2025}
}

@article{jain2025r2egym,
	author       = {Naman Jain and
	Jaskirat Singh and
	Manish Shetty and
	Liang Zheng and
	Koushik Sen and
	Ion Stoica},
	title        = {R2E-Gym: Procedural Environments and Hybrid Verifiers for Scaling
	Open-Weights {SWE} Agents},
	journal      = {arXiv},
	year         = {2025}
}

@inproceedings{yang2025swesmith,
  author       = {John Yang and
                  Kilian Lieret and
                  Carlos E. Jimenez and
                  Alexander Wettig and
                  Kabir Khandpur and
                  Yanzhe Zhang and
                  Binyuan Hui and
                  Ofir Press and
                  Ludwig Schmidt and
                  Diyi Yang},
  title        = {SWE-smith: Scaling Data for Software Engineering Agents},
  booktitle    = {NeurIPS},
  year         = {2025}
}

@article{ahmad2026openswetraces,
  author       = {Wasi Uddin Ahmad and
                  Nikolai Ludwig and
                  Somshubra Majumdar and
                  Boris Ginsburg},
  title        = {Open-SWE-Traces: Advancing Dual-Mode Multilingual Distillation for
                  Software Engineering Agents},
  journal      = {arXiv},
  year         = {2026}
}

@article{tao2026swelego,
  author       = {Chaofan Tao and
                  Jierun Chen and
                  Yuxin Jiang and
                  Kaiqi Kou and
                  Shaowei Wang and
                  Ruoyu Wang and
                  Xiaohui Li and
                  Sidi Yang and
                  Yiming Du and
                  Jianbo Dai and
                  Zhiming Mao and
                  Xinyu Wang and
                  Lifeng Shang and
                  Haoli Bai},
  title        = {SWE-Lego: Pushing the Limits of Supervised Fine-tuning for Software
                  Issue Resolving},
  journal      = {arXiv},
  year         = {2026}
}

@article{ma2026patches,
  author       = {Murong Ma and
                  Tianyu Chen and
                  Yun Lin and
                  Shuai Lu and
                  Qinglin Zhu and
                  Yeyun Gong and
                  Zhiyong Huang and
                  Peng Cheng and
                  Yan Lu and
                  Jin Song Dong},
  title        = {From Patches to Trajectories: Privileged Process Supervision for Software-Engineering
                  Agents},
  journal      = {arXiv},
  year         = {2026}
}

@article{han2026swetrace,
  author       = {Hao Han and
                  Jin Xie and
                  Xuehao Ma and
                  Weiquan Zhu and
                  Ziyao Zhang and
                  ZhiLiang Long and
                  Hongkai Chen and
                  Qingwen Ye},
  title        = {{SWE-TRACE:} Optimizing Long-Horizon {SWE} Agents Through Rubric Process
                  Reward Models and Heuristic Test-Time Scaling},
  journal      = {arXiv},
  year         = {2026}
}

@article{xiao2025agentdiet,
	author       = {Yuan{-}An Xiao and
	Pengfei Gao and
	Chao Peng and
	Yingfei Xiong},
	title        = {Reducing Cost of {LLM} Agents with Trajectory Reduction},
	journal      = {PACMSE},
	year         = {2026}
}

@inproceedings{lin2025seagent,
  author       = {Yifu Guo and
                  Jiaye Lin and
                  Huacan Wang and
                  Yuzhen Han and
                  Sen Hu and
                  Ziyi Ni and
                  Licheng Wang and
                  Mingguang Chen},
  title        = {SE-Agent: Self-Evolution Trajectory Optimization in Multi-Step Reasoning
                  with LLM-Based Agents},
  booktitle    = {NeurIPS},
  year         = {2025}
}

@article{ding2026swereplay,
  author       = {Yifeng Ding and
                  Lingming Zhang},
  title        = {SWE-Replay: Efficient Test-Time Scaling for Software Engineering Agents},
  journal      = {arXiv},
  year         = {2026},
}

@article{kim2026scalingagentic,
	author       = {Joongwon Kim and
	Wannan Yang and
	Kelvin Niu and
	Hongming Zhang and
	Yun Zhu and
	Eryk Helenowski and
	Ruan Silva and
	Zhengxing Chen and
	Srinivasan Iyer and
	Manzil Zaheer and
	Daniel Fried and
	Hannaneh Hajishirzi and
	Sanjeev Arora and
	Gabriel Synnaeve and
	Ruslan Salakhutdinov and
	Anirudh Goyal},
	title        = {Scaling Test-Time Compute for Agentic Coding},
	journal      = {arXiv},
	year         = {2026}
}

@article{ehrlich2025codemonkeys,
  author       = {Ryan Ehrlich and
                  Bradley C. A. Brown and
                  Jordan Juravsky and
                  Ronald Clark and
                  Christopher R{\'{e}} and
                  Azalia Mirhoseini},
  title        = {CodeMonkeys: Scaling Test-Time Compute for Software Engineering},
  journal      = {arXiv},
  year         = {2025}
}

@inproceedings{shinn2023reflexion,
  author       = {Noah Shinn and
                  Federico Cassano and
                  Ashwin Gopinath and
                  Karthik Narasimhan and
                  Shunyu Yao},
  title        = {Reflexion: language agents with verbal reinforcement learning},
  booktitle    = {NeurIPS},
  year         = {2023}
}

@inproceedings{zhao2023expel,
  author       = {Andrew Zhao and
                  Daniel Huang and
                  Quentin Xu and
                  Matthieu Lin and
                  Yong{-}Jin Liu and
                  Gao Huang},
  title        = {ExpeL: {LLM} Agents Are Experiential Learners},
  booktitle    = {AAAI},
  year         = {2024}
}

@inproceedings{wang2024awm,
  author       = {Zora Zhiruo Wang and
                  Jiayuan Mao and
                  Daniel Fried and
                  Graham Neubig},
  title        = {Agent Workflow Memory},
  booktitle    = {ICML},
  year         = {2025}
}

@article{li2026codeskill,
  author       = {Yanzhou Li and
                  Yiran Zhang and
                  Xiaoyu Zhang and
                  Xiaoxia Liu and
                  Yang Liu},
  title        = {{CODESKILL:} Learning Self-Evolving Skills for Coding Agents},
  journal      = {arXiv},
  year         = {2026}
}

@article{chen2025sweexp,
	author       = {Silin Chen and
	Shaoxin Lin and
	Xiaodong Gu and
	Yuling Shi and
	Heng Lian and
	Longfei Yun and
	Dong Chen and
	Weiguo Sun and
	Lin Cao and
	Qianxiang Wang},
	title        = {SWE-Exp: Experience-Driven Software Issue Resolution},
	journal      = {arXiv},
	year         = {2025}
}

@article{lin2024evocoder,
	author       = {Yalan Lin and
	Yingwei Ma and
	Rongyu Cao and
	Binhua Li and
	Fei Huang and
	Xiaodong Gu and
	Yongbin Li},
	title        = {LLMs as Continuous Learners: Improving the Reproduction of Defective
	Code in Software Issues},
	journal      = {arXiv},
	year         = {2024}
}

@article{mu2025experepair,
	author       = {Fangwen Mu and
	Junjie Wang and
	Lin Shi and
	Song Wang and
	Shoubin Li and
	Qing Wang},
	title        = {ExpeRepair: Dual-Memory Enhanced LLM-Based Repository-Level Program
	Repair},
	journal      = {PACMSE},
	year         = {2026}
}

@article{wang2026memgovern,
	author       = {Qihao Wang and
	Ziming Cheng and
	Shuo Zhang and
	Fan Liu and
	Rui Xu and
	Heng Lian and
	Kunyi Wang and
	Xiaoming Yu and
	Jianghao Yin and
	Sen Hu and
	Yue Hu and
	Shaolei Zhang and
	Yanbing Liu and
	Ronghao Chen and
	Huacan Wang},
	title        = {MemGovern: Enhancing Code Agents through Learning from Governed Human
	Experiences},
	journal      = {arXiv},
	year         = {2026}
}

@article{hayashi2025sage,
	author       = {Hiroaki Hayashi and
	Bo Pang and
	Wenting Zhao and
	Ye Liu and
	Akash Gokul and
	Srijan Bansal and
	Caiming Xiong and
	Semih Yavuz and
	Yingbo Zhou},
	title        = {Self-Abstraction from Grounded Experience for Plan-Guided Policy Refinement},
	journal      = {arXiv},
	year         = {2025}
}

@article{xia2025livesweagent,
	author       = {Chunqiu Steven Xia and
	Zhe Wang and
	Yan Yang and
	Yuxiang Wei and
	Lingming Zhang},
	title        = {Live-SWE-agent: Can Software Engineering Agents Self-Evolve on the
	Fly?},
	journal      = {arXiv},
	year         = {2025}
}

@article{xiao2026socraticswe,
  author       = {Chuan Xiao and
                  Zhengbo Jiao and
                  Shaobo Wang and
                  Wei Wang and
                  Bing Zhao and
                  Hu Wei and
                  Linfeng Zhang and
                  Lin Qu},
  title        = {Socratic-SWE: Self-Evolving Coding Agents via Trace-Derived Agent
                  Skills},
  journal      = {arXiv},
  year         = {2026}
}

@inproceedings{madaan2023selfrefine,
  author       = {Aman Madaan and
                  Niket Tandon and
                  Prakhar Gupta and
                  Skyler Hallinan and
                  Luyu Gao and
                  Sarah Wiegreffe and
                  Uri Alon and
                  Nouha Dziri and
                  Shrimai Prabhumoye and
                  Yiming Yang and
                  Shashank Gupta and
                  Bodhisattwa Prasad Majumder and
                  Katherine Hermann and
                  Sean Welleck and
                  Amir Yazdanbakhsh and
                  Peter Clark},
  title        = {Self-Refine: Iterative Refinement with Self-Feedback},
  booktitle    = {NeurIPS},
  year         = {2023}
}

@article{xi2024agentgym,
  author       = {Zhiheng Xi and
                  Yiwen Ding and
                  Wenxiang Chen and
                  Boyang Hong and
                  Honglin Guo and
                  Junzhe Wang and
                  Dingwen Yang and
                  Chenyang Liao and
                  Xin Guo and
                  Wei He and
                  Songyang Gao and
                  Lu Chen and
                  Rui Zheng and
                  Yicheng Zou and
                  Tao Gui and
                  Qi Zhang and
                  Xipeng Qiu and
                  Xuanjing Huang and
                  Zuxuan Wu and
                  Yu{-}Gang Jiang},
  title        = {AgentGym: Evolving Large Language Model-based Agents across Diverse
                  Environments},
  journal      = {arXiv},
  year         = {2024}
}

@article{robeyns2025sica,
  author       = {Maxime Robeyns and
                  Martin Szummer and
                  Laurence Aitchison},
  title        = {A Self-Improving Coding Agent},
  journal      = {arXiv},
  year         = {2025}
}

@article{pan2026rho,
  author       = {Wenbo Pan and
                  Shujie Liu and
                  Chin{-}Yew Lin and
                  Jingying Zeng and
                  Xianfeng Tang and
                  Xiangyang Zhou and
                  Yan Lu and
                  Xiaohua Jia},
  title        = {Evolving Agents in the Dark: Retrospective Harness Optimization via
                  Self-Preference},
  journal      = {arXiv},
  year         = {2026}
}

@article{lin2026ahe,
  author       = {Jiahang Lin and
                  Shichun Liu and
                  Chengjun Pan and
                  Lizhi Lin and
                  Shihan Dou and
                  Xuanjing Huang and
                  Hang Yan and
                  Zhenhua Han and
                  Tao Gui},
  title        = {Agentic Harness Engineering: Observability-Driven Automatic Evolution
                  of Coding-Agent Harnesses},
  journal      = {arXiv},
  year         = {2026}
}

@inproceedings{sener2018coreset,
	author       = {Ozan Sener and
	Silvio Savarese},
	title        = {Active Learning for Convolutional Neural Networks: {A} Core-Set Approach},
	booktitle    = {ICLR},
	year         = {2018}
}

@inproceedings{kirsch2019batchbald,
	author       = {Andreas Kirsch and
	Joost van Amersfoort and
	Yarin Gal},
	title        = {BatchBALD: Efficient and Diverse Batch Acquisition for Deep Bayesian
	Active Learning},
	booktitle    = {NeurIPS},
	year         = {2019}
}

@inproceedings{foster2021dad,
	author       = {Adam Foster and
	Desi R. Ivanova and
	Ilyas Malik and
	Tom Rainforth},
	title        = {Deep Adaptive Design: Amortizing Sequential Bayesian Experimental
	Design},
	booktitle    = {ICML},
	year         = {2021}
}

@inproceedings{koh2017influence,
	author       = {Pang Wei Koh and
	Percy Liang},
	title        = {Understanding Black-box Predictions via Influence Functions},
	booktitle    = {ICML},
	year         = {2017}
}

@inproceedings{ghorbani2019datashapley,
	author       = {Amirata Ghorbani and
	James Y. Zou},
	title        = {Data Shapley: Equitable Valuation of Data for Machine Learning},
	booktitle    = {ICML},
	year         = {2019}
}

@inproceedings{paul2021datadiet,
	author       = {Mansheej Paul and
	Surya Ganguli and
	Gintare Karolina Dziugaite},
	title        = {Deep Learning on a Data Diet: Finding Important Examples Early in
	Training},
	booktitle    = {NeurIPS},
	year         = {2021}
}

@inproceedings{xie2023doremi,
	author       = {Sang Michael Xie and
	Hieu Pham and
	Xuanyi Dong and
	Nan Du and
	Hanxiao Liu and
	Yifeng Lu and
	Percy Liang and
	Quoc V. Le and
	Tengyu Ma and
	Adams Wei Yu},
	title        = {DoReMi: Optimizing Data Mixtures Speeds Up Language Model Pretraining},
	booktitle    = {NeurIPS},
	year         = {2023}
}

@inproceedings{bellemare2016pseudocount,
	author       = {Marc G. Bellemare and
	Sriram Srinivasan and
	Georg Ostrovski and
	Tom Schaul and
	David Saxton and
	R{\'{e}}mi Munos},
	title        = {Unifying Count-Based Exploration and Intrinsic Motivation},
	booktitle    = {NeurIPS},
	year         = {2016}
}

@inproceedings{pathak2017curiosity,
	author       = {Deepak Pathak and
	Pulkit Agrawal and
	Alexei A. Efros and
	Trevor Darrell},
	title        = {Curiosity-driven Exploration by Self-supervised Prediction},
	booktitle    = {ICML},
	year         = {2017}
}

@article{gideoni2026simple,
	author       = {Yonatan Gideoni and
	Sebastian Risi and
	Yarin Gal},
	title        = {Simple Baselines are Competitive with Code Evolution},
	journal      = {arXiv},
	year         = {2026}
}

@inproceedings{kocsis2006bandit,
	author       = {Levente Kocsis and
	Csaba Szepesv{\'{a}}ri},
	title        = {Bandit Based Monte-Carlo Planning},
	booktitle    = {ECML},
	year         = {2006}
}

@inproceedings{zhou2023lats,
	author       = {Andy Zhou and
	Kai Yan and
	Michal Shlapentokh{-}Rothman and
	Haohan Wang and
	Yu{-}Xiong Wang},
	title        = {Language Agent Tree Search Unifies Reasoning, Acting, and Planning
	in Language Models},
	booktitle    = {ICML},
	year         = {2024}
}

@inproceedings{antoniades2024swesearch,
	author       = {Antonis Antoniades and
	Albert {\"{O}}rwall and
	Kexun Zhang and
	Yuxi Xie and
	Anirudh Goyal and
	William Yang Wang},
	title        = {SWE-Search: Enhancing Software Agents with Monte Carlo Tree Search
	and Iterative Refinement},
	booktitle    = {ICLR},
	year         = {2025}
}

@inproceedings{chen2026tdscaling,
	author       = {Guhong Chen and
	Chenghao Sun and
	Cheng Fu and
	Qiyao Wang and
	Zhihong Huang and
	Chaopeng Wei and
	Guangxu Chen and
	Feiteng Fang and
	Ahmadreza Argha and
	Bing Zhao and
	Xander Xu and
	Qi Han and
	Hamid Alinejad{-}Rokny and
	Qiang Qu and
	Binhua Li and
	Shiwen Ni and
	Min Yang and
	{HU} Wei and
	Yongbin Li},
	title        = {Beyond Quantity: Trajectory Diversity Scaling for Code Agents},
	booktitle    = {ACL},
	year         = {2026}
}

@inproceedings{han2026trajectorycuration,
	author       = {Yunze Han},
	title        = {A Systematic Evaluation of Trajectory Data Curation for LoRA Fine-Tuning
	of Code Agents},
	booktitle    = {ICIC},
	year         = {2026}
}

@article{wang2026curateevo,
	author       = {Dingzirui Wang and
	Xuanliang Zhang and
	Keyan Xu and
	Qingfu Zhu and
	Wanxiang Che},
	title        = {CurateEvo: Data-Curation Evolving for Agentic Post-Training},
	journal      = {arXiv},
	year         = {2026}
}

@misc{inclusionai2026ling3flashbase,
  author       = {{inclusionAI}},
  title        = {{Ling-3.0-flash-base}},
  year         = {2026},
  howpublished = {Hugging Face model card},
  url          = {https://huggingface.co/inclusionAI/Ling-3.0-flash-base}
}

@article{qwenteam2025qwen3base,
	author       = {{Qwen Team}},
	title        = {Qwen3 Technical Report},
	journal      = {arXiv},
	year         = {2025}
}

@article{badertdinov2026swerebenchv2,
	author       = {Ibragim Badertdinov and
	Maksim Nekrashevich and
	Anton Shevtsov and
	Alexander Golubev},
	title        = {SWE-rebench {V2:} Language-Agnostic {SWE} Task Collection at Scale},
	journal      = {arXiv},
	year         = {2026}
}

@article{glm5team2026glm5,
	title   = {{GLM-5}: From Vibe Coding to Agentic Engineering},
	author  = {{GLM-5 Team}},
	journal = {arXiv},
	year    = {2026}
}

@misc{zai2026glm51,
	title        = {{GLM-5.1}},
	author       = {{Z.ai}},
	year         = {2026},
	howpublished = {Hugging Face model card},
	url          = {https://huggingface.co/zai-org/GLM-5.1}
}

@misc{openai2026gpt56,
	title        = {{GPT-5.6} System Card},
	author       = {{OpenAI}},
	year         = {2026},
	howpublished = {Technical report},
	url          = {https://deploymentsafety.openai.com/gpt-5-6},
	note         = {Accessed August 2026}
}

@article{deng2025swebenchpro,
	author       = {Xiang Deng and
	Jeff Da and
	Edwin Pan and
	Yannis Yiming He and
	Charles Ide and
	Kanak Garg and
	Niklas Lauffer and
	Andrew Park and
	Nitin Pasari and
	Chetan Rane and
	Karmini Sampath and
	Maya Krishnan and
	Srivatsa Kundurthy and
	Sean Hendryx and
	Zifan Wang and
	Chen Bo Calvin Zhang and
	Noah Jacobson and
	Bing Liu and
	Brad Kenstler},
	title        = {SWE-Bench Pro: Can {AI} Agents Solve Long-Horizon Software Engineering
	Tasks?},
	journal      = {arXiv},
	year         = {2025}
}

@article{sun2026judging,
	title={Judging the Justification: Decision Certificate Supervision for Generative Reward Models},
	author={Sun, Guiquan and Bai, Bing and Min, Martin Renqiang and Song, Dongjin},
	journal={arXiv preprint arXiv:2610.27528},
	year={2026}
}

@article{li2026same,
	title={Same Signal, Opposite Meaning: Direction-Informed Adaptive Learning for LLM Agents},
	author={Li, Ziming and Huang, Jiatan and Guo, Xiaoguang and Wang, Guilin and Zhang, Chuxu},
	journal={arXiv preprint arXiv:2605.06908},
	year={2026}
}

@article{huang2026cleat,
	title={CLEAT: Learning Value-Guided Clarification for Interactive Agents},
	author={Huang, Jiatan and Zhang, Zheyuan and Li, Ziming and Ma, Tianyi and Li, Mingchen and Ye, Yanfang and Zhang, Chuxu},
	journal={arXiv preprint arXiv:2610.12655},
	year={2026}
}

@article{ma2026proplay,
	title={ProPlay: Procedural World Models for Self-Evolving LLM Agents},
	author={Ma, Yijun and Wang, Zehong and Li, Yiyang and Li, Ziming and Guo, Xiaoguang and Sun, Weixiang and Zhang, Chuxu and Ye, Yanfang},
	journal={arXiv preprint arXiv:2606.12780},
	year={2026}
}
